\documentclass[journal]{IEEEtran}

\ifCLASSINFOpdf
\else
\fi

\usepackage{times}
\usepackage{epsfig}
\usepackage{graphicx}
\usepackage{amsmath}
\usepackage{amssymb}
\usepackage{mathrsfs}
\usepackage{multirow}
\usepackage{blkarray}
\usepackage{mathtools}
\usepackage{float}
\usepackage{array}
\usepackage{epstopdf}
\usepackage{subcaption}
\usepackage[T1]{fontenc}
\usepackage{algorithm}% http://ctan.org/pkg/algorithm
\usepackage{algpseudocode}

\begin{document}

\title{Screen Content Image Segmentation Using Sparse Decomposition and Total Variation Minimization}

\author{Shervin~Minaee,~\IEEEmembership{Student Member,~IEEE,}
        and~Yao~Wang,~\IEEEmembership{Fellow,~IEEE}% <-this % stops a space
%\thanks{M. Shell is with the Department
%of Electrical and Computer Engineering, Georgia Institute of Technology, Atlanta,
%GA, 30332 USA e-mail: (see http://www.michaelshell.org/contact.html).}% <-this % stops a space
%\thanks{J. Doe and J. Doe are with Anonymous University.}% <-this % stops a space
%\thanks{Manuscript received April 19, 2005; revised September 17, 2014.}
}

% The paper headers
%\markboth{Journal of \LaTeX\ Class Files,~Vol.~13, No.~9, September~2014}%
%{Shell \MakeLowercase{\textit{et al.}}: Background/Foreground Segmentation Using Robust Regression Approach}

\maketitle

% As a general rule, do not put math, special symbols or citations
% in the abstract or keywords.
\begin{abstract}
Sparse decomposition has been widely used for different applications, such as source separation, image classification, image denoising and more. 
This paper presents a new algorithm for segmentation of an image into background and foreground text and graphics using sparse decomposition and total variation minimization.
The proposed method is designed based on the assumption that the background part of the image is smoothly varying and can be represented by a linear combination of a few smoothly varying basis functions, while the foreground text and graphics can be modeled with a sparse component overlaid on the smooth background.
The background and foreground are separated using a sparse decomposition framework regularized with a few suitable regularization terms which promotes the sparsity and connectivity of foreground pixels.
This algorithm has been tested on a dataset of images extracted from HEVC standard test sequences for screen content coding, and is shown to have superior performance over some prior methods, including least absolute deviation fitting, k-means clustering based segmentation in DjVu and shape primitive extraction and coding (SPEC) algorithm. 
%This segmentation algorithm can be used in different applications such as text extraction, separate coding of background and foreground for compression of screen content and mixed content documents, principal line extraction from palmprint images and medical image segmentation.
\end{abstract}

%\begin{IEEEkeywords}
%Image segmentation, sparse decomposition, total variation, ADMM.
%\end{IEEEkeywords}

\IEEEpeerreviewmaketitle

\section{Introduction}
Decomposition of an image into multiple components has many applications in different tasks. One special case is the background-foreground segmentation, which tries to decompose an image into two components, background and foreground.
It has many applications in image processing such as separate coding of background and foreground for compression of mixed content images \cite{MRC}, denoising, medical image segmentation \cite{chromosome}, text extraction \cite{text}, and pre-processing steps for biometrics recognition \cite{bio1},\cite{bio2}.

Different algorithms have been proposed in the past for foreground-background segmentation in still images such as hierarchical k-means clustering in DjVu \cite{djvu}, shape primitive extraction and coding (SPEC) \cite{spec} and least absolute deviation fitting \cite{LAD}.

The hierarchical k-means clustering applies the k-means clustering algorithm with k=2 on blocks in multi-resolution. It first applies the k-means clustering algorithm on a large block to obtain foreground and background colors and then uses them as the initial foreground and background colors for the smaller blocks in the next stages. It also applies some post-processing at the end to refine the results. This algorithm has difficulty for the regions where background and foreground color intensities overlap. %and some part of the background will be detected as foreground or the other way.
In SPEC algorithm, a two-step segmentation algorithm is proposed. 
In the first step the algorithm classifies each block of size $16 \times 16$ into either pictorial block or text/graphics, by comparing the number of colors with the threshold 32.
In the second step, it refines the segmentation result of pictorial blocks, by extracting shape primitives. 
% and then comparing the size and color of the shape primitives with some threshold.
Because blocks containing smoothly varying background over a narrow range can also have a small color number,  it is hard to find a fixed color number threshold that can robustly separate pictorial blocks and text/graphics blocks. 
%Furthermore, text and graphics  in screen content images typically have some variation in their colors, even in the absence of sub-pixel rendering. 
We have previously proposed a least absolute deviation fitting method, which fits a smooth model to the image and classifies the pixels to either background or foreground based on the fitting error \cite{LAD}. It use the $\ell_1$ norm on the fitting error to enforce the sparsity of the error term. Although this algorithm achieved significantly better segmentation than both DjVu and SPEC, it suffers from the existence of isolated points in foreground.
% Most of the previous works regarding foreground segmentation are based on color clustering or color counting and have difficulty for cases where the background color has a large dynamic range or is similar to the foreground color in some regions.

The issues with previous algorithms motivate us to design a new segmentation algorithm which overcomes the problems of previous algorithms. 
We propose a sparse-decomposition framework to perform this image segmentation task.
Sparse representation has been used for various applications in recent years, including face recognition \cite{wright}, visual tracking \cite{taalimi}, morphological component analysis \cite{donoho}, recognition \cite{rahimi1}, image decomposition \cite{starck}, \cite{sherv_asilomar} and image restoration \cite{mairal}.
Despite the huge application of sparse representation, there has not been many works using sparsity for still image segmentation. 

Instead of looking at the intensities of pixels and deciding whether it should belong to background or foreground, we believe it is better to look at the smoothness of a group of pixels and then decide whether they should belong to background or foreground. The other important observation is that, the foreground layer should contain a set of connected pixels (such as the pixels in a text stroke, or a line in graphics), not a set of randomly located isolated points. Therefore in our segmentation algorithm, we enforce the extracted foreground pixels to be connected to each other by penalizing their total variation. 
Based on these two notions, we propose a sparse decomposition framework for the segmentation task.
We model the background part of the image with a linear combination of a set of smoothly varying basis functions, and the foreground layer with a sparse component that have connected pixels.. 
%By penalizing the total variation of the foreground layer we can encourage the foreground pixels to be connected.
A problem with our prior least absolute deviation approach is that it uses a fixed set of smooth basis functions and does not impose any prior on the weighting coefficients. It was challenging to determining the right set of smooth bases that can represent all background patterns well. When too few bases are used, some part of a complicated background will be considered foreground;  On the other hand, with too many bases, some foreground pixels can be fitted into the smooth model, hence falsely segmented as the background. Our new formulation overcomes this difficulty by using a relatively rich set of bases, but enforcing the resulting coefficients to be sparse, by adding a regularization term, which penalizes the use of many bases for background representation. The proposed algorithm here, can also be used for decomposition of other types of signals \cite{rahimi2}.

The structure of the rest of this paper is as follows: Section II presents the core idea of the proposed segmentation methods. Section III describes the ADMM formulation for solving the sparse decomposition based segmentation. Section IV provides the experimental results for the proposed algorithm. And finally the paper is concluded in Section V.

\section{Sparse Decomposition Framework}
It is clear that smooth background regions can be well represented with a few smooth basis functions, 
%By well representation we mean that the approximated value at a pixel with the smooth functions should have an error less than a desired threshold at every pixel. 
whereas the high-frequency component of the image belonging to the foreground, cannot be modeled with a smooth model. But using the fact that foreground pixels occupy a relatively small percentage of the images we can model them with a sparse component overlaid on background. Therefore it is fairly natural to think of mixed content image as a superposition of two components, one smooth and the other one sparse. Therefore we can use signal decomposition techniques to separate these two components.

%Sparsity-based signal decomposition has drawn a lot of attention in recent years because of its tremendous success in achieving  state-of-the-art results in many areas.
%To decompose a signal $x$ into its components as $x= x_1+x_2$ using sparse representation, usually two or more suitable dictionaries are found which are suitable for sparse representation of only one signal component, but not the others \cite{starck}.

We first need to derive a suitable model for background component. We divide each image into non-overlapping blocks of size $N\times N$, and then represent each image block denoted by $F(x,y)$, with a smooth model $B(x,y;\alpha_1,...,\alpha_K)$, where $x$ and $y$ denote the horizontal and vertical axes and $\alpha_1,...,\alpha_K$ denote the parameters of this smooth model.
For the choice of smooth model, we propose to use a linear combination of $K$ basis functions $\sum_{k=1}^K \alpha_k P_k(x,y)$, where $P_k(x,y)$ denotes a 2D smooth basis function \cite{LAD}. 
%Since this model is a linear function of parameters, $\alpha_k$, it is simpler to find the optimal weights.
We applied Karhunen-Loeve transform \cite{KLT} to a training set of smooth background images and the optimal transforms turns out to be very similar to 2D DCT bases.
Therefore we used a set of low frequency two-dimensional DCT basis functions, since they have been shown to be very efficient for image representation \cite{DCT}. The 2-D DCT function is defined as:
\begin{equation*}
P_{u,v}(x,y)= \beta_u \beta_v cos((2x+1)\pi u/2N) cos((2y+1)\pi v/2N) 
\end{equation*}
where $u$ and $v$ denote the frequency of the basis and $\beta_u$ and $\beta_v$ are normalization factors.
We order all the possible basis functions in the conventional zig-zag order in the $(u,v)$ plane, and choose the first $K$ basis functions.
Since we do not know in advance how many basis functions to include for the background part, we allow the model to choose from a large set of bases that we think are sufficient to represent the most "complex" background, while minimizing coefficient $\ell_0$ norm. Without such a restriction on the coefficients, we might end up with the situation that all foreground pixels are also modeled by the smooth layer.

Overall, each image block is represented as:
\begin{equation}
F(x,y)= \sum_{k=1}^K \alpha_k P_k(x,y) + S(x,y)
\end{equation}
where $\sum_{i=1}^K \alpha_i P_i(x,y)$ and $S(x,y)$ correspond to the smooth background region and foreground pixels respectively. So after decomposition, those pixels with large value in the $S$ component will be considered as foreground.

To have a more compact notation, we can look at the 1D version of this problem by converting the 2D blocks of size $N \times N$ into a vector of length $N^2$, denoted by $f$, and denoting $\sum_{k=1}^K \alpha_k P_k(x,y)$ as $ {P\alpha}$ where ${P}$ is a matrix of size $N^2\times K$ in which the $k$-th column corresponds to the vectorized version of $P_k(x,y)$ and $\alpha=[\alpha_1,...,\alpha_K]^\text{T}$. The 1D version of $S(x,y)$ is denoted by $s$. 
Then Eq. (1) can be written as:
\begin{equation}
f= {P\alpha}+s
\end{equation}

Now to perform image segmentation, we need to impose some prior knowledge about background and foreground to our optimization problem. First of all, as described earlier, we do not want to use too many basis functions for background representation, since by using many basis for background we might be able to even represent some part of the foreground regions with the smooth model and consider them as background (imagine the case that we use a complete set of bases for background representation). Therefore the number of nonzero components of $\alpha$ should be small ( i.e. $\| \alpha \|_0$ should be small).
On the other hand we expect the majority of the pixels in each block to belong to the background component, therefore the number of nonzero components of $s$ should not be very large. This feature is very desirable in image and video compression applications, because the background component can be easily represented using a set of low-order DCT bases. So the more background pixels we have, the smaller bit-rate we usually need.
And the last but not the least point is that we expect the nonzero component of the foreground to be connected to each other, therefore we can add a regularization term which promotes the connectivity of foreground pixels. Here we used total variation of the foreground component to penalize isolated points in foreground.
Putting all of these priors together we will get the following optimization problem:
\begin{equation}
\begin{aligned}
& \underset{s, \alpha}{\text{minimize}}
& & \|\alpha \|_0+ \lambda_1 \| s \|_0+ \lambda_2 TV(s)   \\
& \text{subject to}
& &  f=P \alpha+s
\end{aligned}
\end{equation}
where $\lambda_1$ and $\lambda_2$ are some constants that need to be tuned. 
For the first two terms since $\ell_0$ is not convex, we use its approximated $\ell_1$ version to have a convex problem.
For the total variation we can use either the isotropic or the anisotropic version of 2D total variation \cite{TV}. To make our optimization problem simpler, we have used the anisotropic version in this algorithm, which is defined as:
\begin{equation}
\begin{aligned}
TV(s)=  \sum_{i,j} |S_{i+1,j}-S_{i,j}|+|S_{i,j+1}-S_{i,j}|
\end{aligned}
\end{equation}
After converting the 2D blocks into 1D vector, we can denote the total variation as below:
\begin{equation}
\begin{aligned}
TV(s)= \| D_xs \|_1+\| D_ys \|_1= \|Ds\|_1
\end{aligned}
\end{equation}
where $D_x$ and $D_y$ are the horizontal and vertical gradient operator matrices, and $D=[ D_x',D_y']'$. Then we will get the following problem:
\begin{equation}
\begin{aligned}
& \underset{s, \alpha}{\text{minimize}}
& & \|\alpha \|_1+ \lambda_1 \| s \|_1+ \lambda_2 \|Ds\|_1   \\
& \text{subject to}
& &  P \alpha+s=f
\end{aligned}
\end{equation}
From the constraint in the above problem, we get $s= f-P\alpha$ and then we derive the following unconstrained problem:
\begin{equation}
\begin{aligned}
& \underset{ \alpha}{\text{min}}
& & \|\alpha \|_1+ \lambda_1 \| f-P\alpha \|_1+ \lambda_2 \|Df-DP\alpha\|_1
\end{aligned}
\end{equation}
This problem can be solved with different approaches, such as alternating direction method of multipliers (ADMM) \cite{ADMM}, majorization minimization \cite{seles1} and forward-backward-forward algorithm (FBF) \cite{patrick1}. Here we present the formulation using ADMM algorithm.

\section{ADMM formulation for the proposed sparse decomposition}
ADMM (Alternating Direction Method of Multipliers) is a popular algorithm which combines superior convergence properties of method of multiplier and decomposability of dual ascent. %This algorithm is also known under other names such as
The ADMM formulation for the optimization problem in (7) can be derived as:
\begin{equation}
\begin{aligned}
& \underset{\alpha, y, z, x}{\text{minimize}}
& & \|y \|_1+ \lambda_1 \| z \|_1+ \lambda_2 \|x\|_1   \\
& \text{subject to} & &  y=\alpha \\
& & & z=f-P \alpha \\
& & & x=Df-DP\alpha
\end{aligned}
\end{equation}
Then the augmented Lagrangian for the above problem can be formed as:
\begin{align*}
&L_{\rho1,\rho2,\rho3}(\alpha,y,z,x)= \|y \|_1+ \lambda_1 \| z \|_1+ \lambda_2 \|x\|_1+ u_1^t(y- \alpha)+ \nonumber \\
&u_2^t(z+P\alpha-f)+ u_3^t(x+DP\alpha-Df)+\frac{\rho_1}{2} \| y- \alpha \|_2^2+  \nonumber \\
&\frac{\rho_2}{2} \| z+P\alpha-f \|_2^2+\frac{\rho_3}{2} \| x+DP\alpha-Df \|_2^2
\end{align*}
where $u_{1:3}$ and $\rho_{1:3}$ denote the dual variables and penalty parameters respectively.
Then, by taking the gradient of the objective function w.r.t. to the primal variables and setting it to zero and using dual descent for dual variables, we will get update rule described in Algorithm 1:

\begin{algorithm}
  \caption{pseudo-code for ADMM updates of problem (8)}\label{euclid}
  \begin{algorithmic}[1]
      \For{\texttt{$k$=1:$k_{max}$}}
        \State $\alpha^{k+1}= A^{-1} \big[ u_1^{k}-P^tu_2^{k}-P^tD^tu_3^{k}+\rho_1y^k$
        \Statex \qquad \qquad $ +\rho_2P^t(f-z^{k})+\rho_3P^tD^t(Df-x^{k}) \big] $
        \State $y^{k+1}= \text{Soft}(\alpha^k- \frac{1}{\rho_1} u_1^{k},\frac{1}{\rho_1}) $        
        \State $z^{k+1}= \text{Soft}(f-P\alpha^{k+1}-\frac{1}{\rho_2} u_2^{k},\frac{\lambda_1}{\rho_2}) $        
        \State $x^{k+1}= \text{Soft}(Df-DP\alpha^{k+1}-\frac{1}{\rho_3} u_3^{k},\frac{\lambda_2}{\rho_3}) $        
        \State $u_1^{k+1}= u_1^{k}+ \rho_1 (y^{k+1}-\alpha^{k+1}) $        
        \State $u_2^{k+1}= u_2^{k}+ \rho_2 (z^{k+1}+P\alpha^{k+1}-f) $
        \State $u_3^{k+1}= u_3^{k}+ \rho_3 (x^{k+1}+DP\alpha^{k+1}-Df) $                
      \EndFor
  \end{algorithmic}
\end{algorithm}

where $A=(\rho_3 P^tD^tDP+\rho_2P^tP+\rho_1 I)$, and $\text{Soft}(x,\lambda)$ denotes the soft-thresholding operator applied elementwise and is defined as:
\begin{gather*}
\text{Soft}(x,\lambda)= \text{sign}(x) \ \text{max}(|x|-\lambda,0)
\end{gather*}

%We can further simplify these updates  which results in the following updates:
%\begin{algorithm}
%  \caption{pseudo-code for ADMM updates of problem (20)}\label{euclid}
%  \begin{algorithmic}[1]
%      \For{\texttt{$k$=1:$k_{max}$}}
%        \State $\alpha^{k+1}= \underset{\alpha}{\text{argmin}} \ L_{\rho1,\rho2,\rho3}(\alpha,y^k,z^k,x^k, u_1^k, u_2^k, u_3^k)$
%        \State $y^{k+1}= \underset{y}{\text{argmin}} \ L_{\rho1,\rho2,\rho3}(\alpha^{k+1},y,z^k,x^k, u_1^k, u_2^k, u_3^k)$        
%        \State $z^{k+1}= \underset{z}{\text{argmin}} \ L_{\rho1,\rho2,\rho3}(\alpha^{k+1},y^{k+1},z,x^k, u_1^k, u_2^k, u_3^k)$        
%        \State $x^{k+1}= \underset{x}{\text{argmin}} \ L_{\rho1,\rho2,\rho3}(\alpha^{k+1},y^{k+1},z^{k+1},x, u_1^k, u_2^k, u_3^k)$        
%        \State $u_1^{k+1}= u_1^{k}+ \rho_1 (y^{k+1}-\alpha^{k+1})$        
%        \State $u_2^{k+1}= u_2^{k}+ \rho_2 (z^{k+1}+P\alpha^{k+1}-f)$
%        \State $u_3^{k+1}= u_3^{k}+ \rho_3 (x^{k+1}+DP\alpha^{k+1}-Df)$                
%      \EndFor
%  \end{algorithmic}
%\end{algorithm}

\section{Experimental Results}
To enable rigorous evaluation of different algorithms, we have generated an extended version of dataset in \cite{LAD}, consisting of 332 image blocks of size 64x64, extracted from sample frames from HEVC test sequences for screen content coding \cite{SCC_data}. The ground truth foregrounds for these images are extracted manually by the author and then refined independently by another person. This dataset is publicly available at \cite{our_dataset}.

In our implementation, the block size is chosen to be $N$=64, which is the same as the largest CU size in HEVC standard. The number of DCT basis functions, $K$, is chosen to be 20. The weight parameters in the objective function are tuned by testing on a validation set (consist of 70 blocks of 64x64) and are set to be $\lambda_1=10$ and $\lambda_2=4$. 
The ADMM algorithm described in Algorithm 1 is implemented in MATLAB and we made it publicly available in \cite{our_dataset}.
The number of iteration for ADMM is chosen to be 50 and the parameter $\rho_1$, $\rho_2$ and $\rho_3$ are all set to 1 as in \cite{boyd}.

We compare the proposed algorithm with three previous algorithms; hierarchical k-means clustering in DjVu, SPEC and least absolute deviation fitting. For SPEC, we have adapted the color number threshold and the shape primitive size threshold from the default value given in \cite{spec}  when necessary to give more satisfactory result. Furthermore, for blocks classified as text/graphics based on the color number, we segment the most frequent color and any similar color to it (i.e. colors whose distance from most frequent color is less than 10 in luminance) in the current block as background and the rest as foreground.

To provide a numerical comparison between the proposed scheme and previous approaches, we have calculated the average precision and recall and F1 score (AKA F-measure) achieved by different segmentation algorithms over this dataset. The average precision, recall and F1 score by different algorithms are given in Table 1. 

The precision and recall are defined as in Eq. (10), where TP, FP and FN denote true positive, false positive and false negative respectively. In our evaluation, we treat a foreground pixel as positive. A pixel that is correctly identified as foreground (compared to the manual segmentation) is considered true positive. The same holds for false negative and false positive. 
\begin{gather}
 \text{Precision}= \frac{\text{TP}}{\text{TP+FP}} \ , 
\ \ \ \ \text{Recall}= \frac{\text{TP}}{\text{TP+FN}} %%%% True Positive Rate 
\end{gather}
The balanced F1 score is defined as the harmonic mean of precision and recall.
\begin{gather*}
\text{F1}= 2 \ \frac{\text{precision} \times \text{recall}}{\text{precision+recall}}
\end{gather*}

As can be seen, the proposed scheme achieved much higher precision and recall than hierarchical k-means and SPEC algorithms. Compared to the least absolute deviation fitting, the proposed formulation yields significant improvement in terms of precision, while also having a slightly higher recall rate.

To see the visual quality of the segmentation, the results for 5 test images (each consisting of multiple 64x64 blocks) are shown in Fig. 1.

\begin{figure*}[ht]
        \centering
        \vspace{-0.5cm}
        \begin{subfigure}[b]{0.18\textwidth}
       % \hspace{-1cm}
                \includegraphics[width=\textwidth]{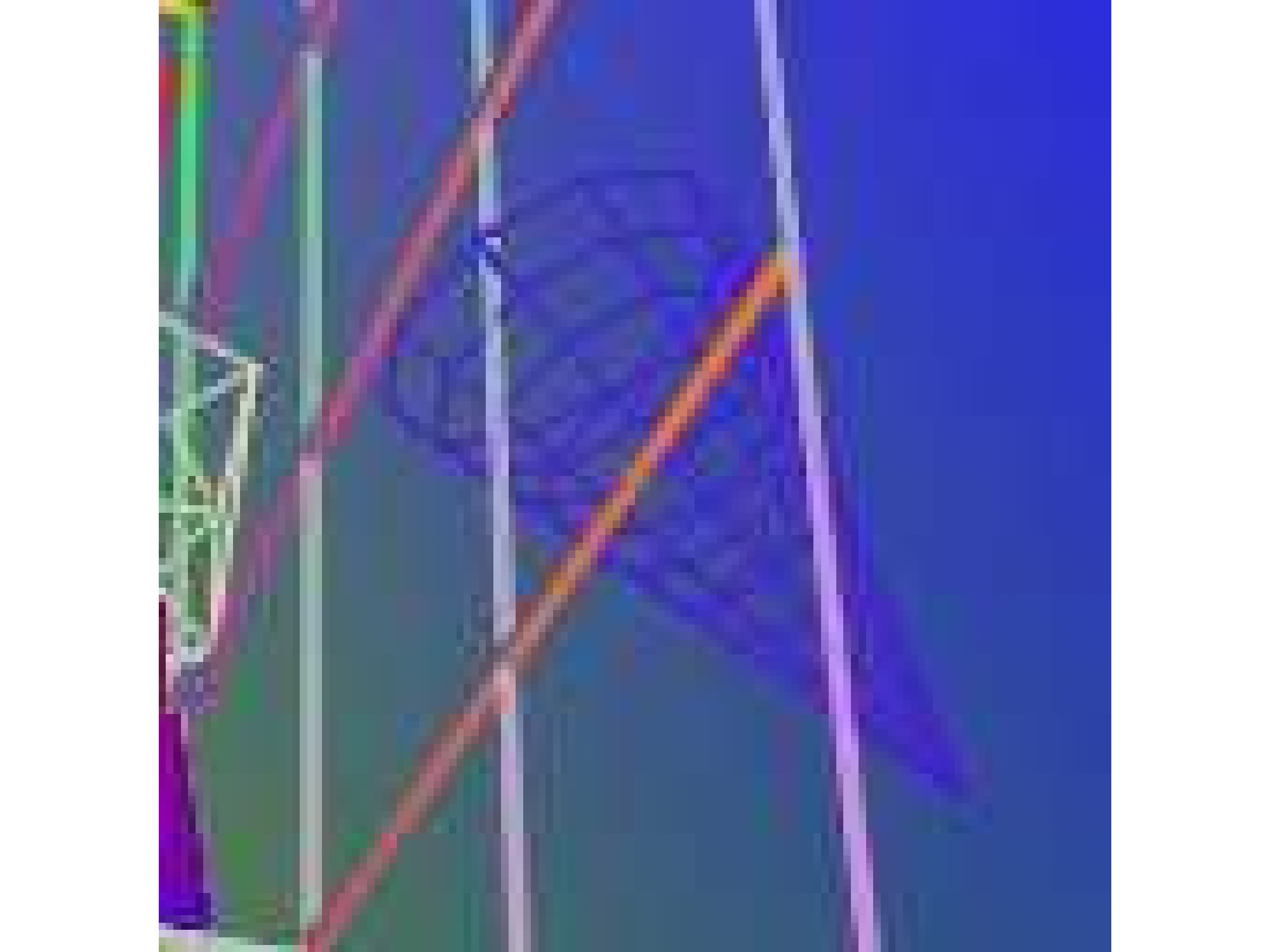}
                                \vspace{-0.5cm}
          \hspace{-2.5cm}    
        \end{subfigure}%
        ~ %add desired spacing between images, e. g. ~, \quad, \qquad, \hfill etc.
          %(or a blank line to force the subfigure onto a new line)
        \begin{subfigure}[b]{0.18\textwidth}
       % \hspace{-2cm}
                \includegraphics[width=\textwidth]{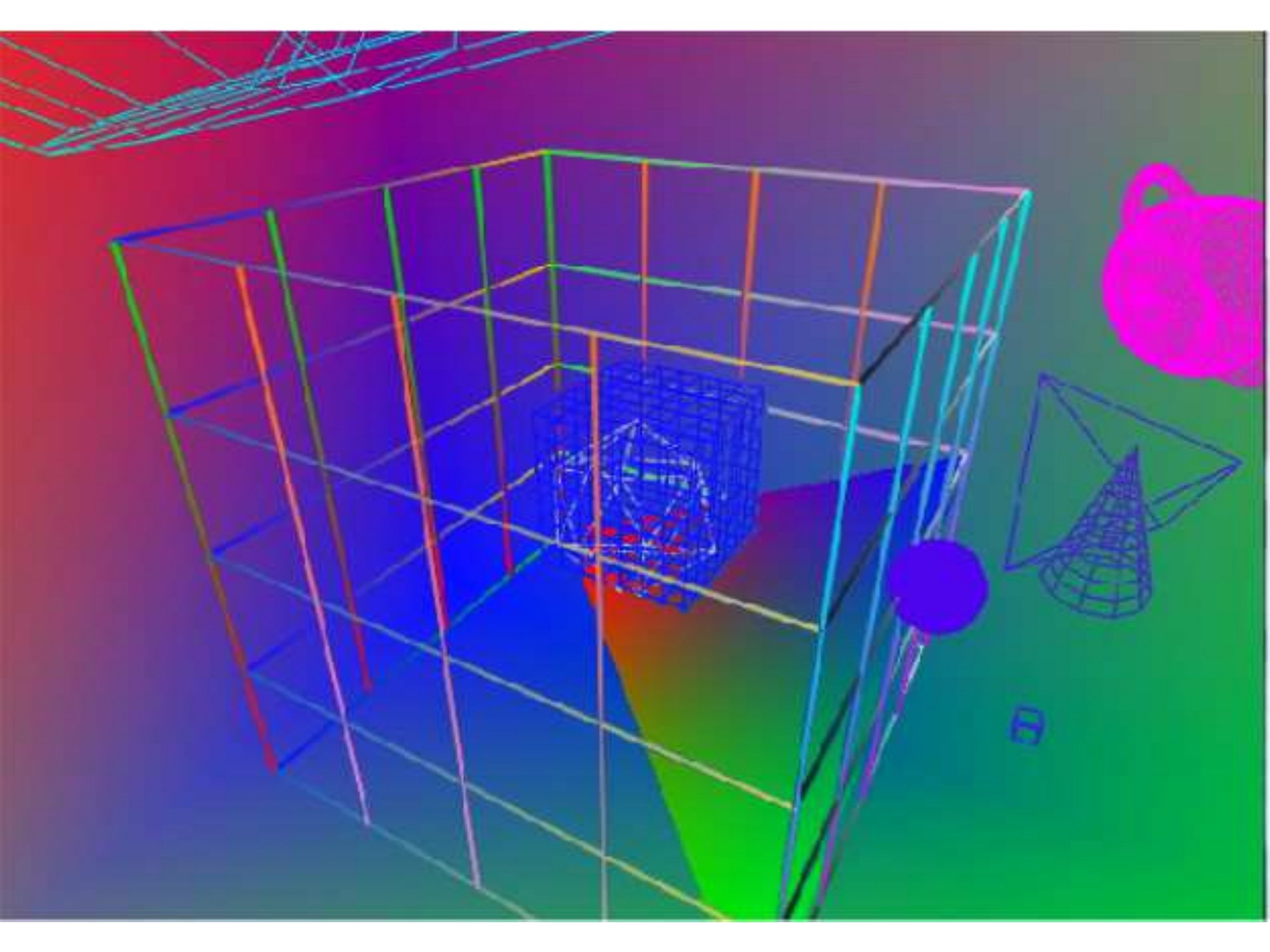}
                \vspace{-0.5cm}
            \hspace{-3cm} 
        \end{subfigure}%
        ~ %add desired spacing between images, e. g. ~, \quad, \qquad, \hfill etc.
          %(or a blank line to force the subfigure onto a new line)
        \begin{subfigure}[b]{0.18\textwidth}
       % \hspace{-2cm}
                \includegraphics[width=\textwidth]{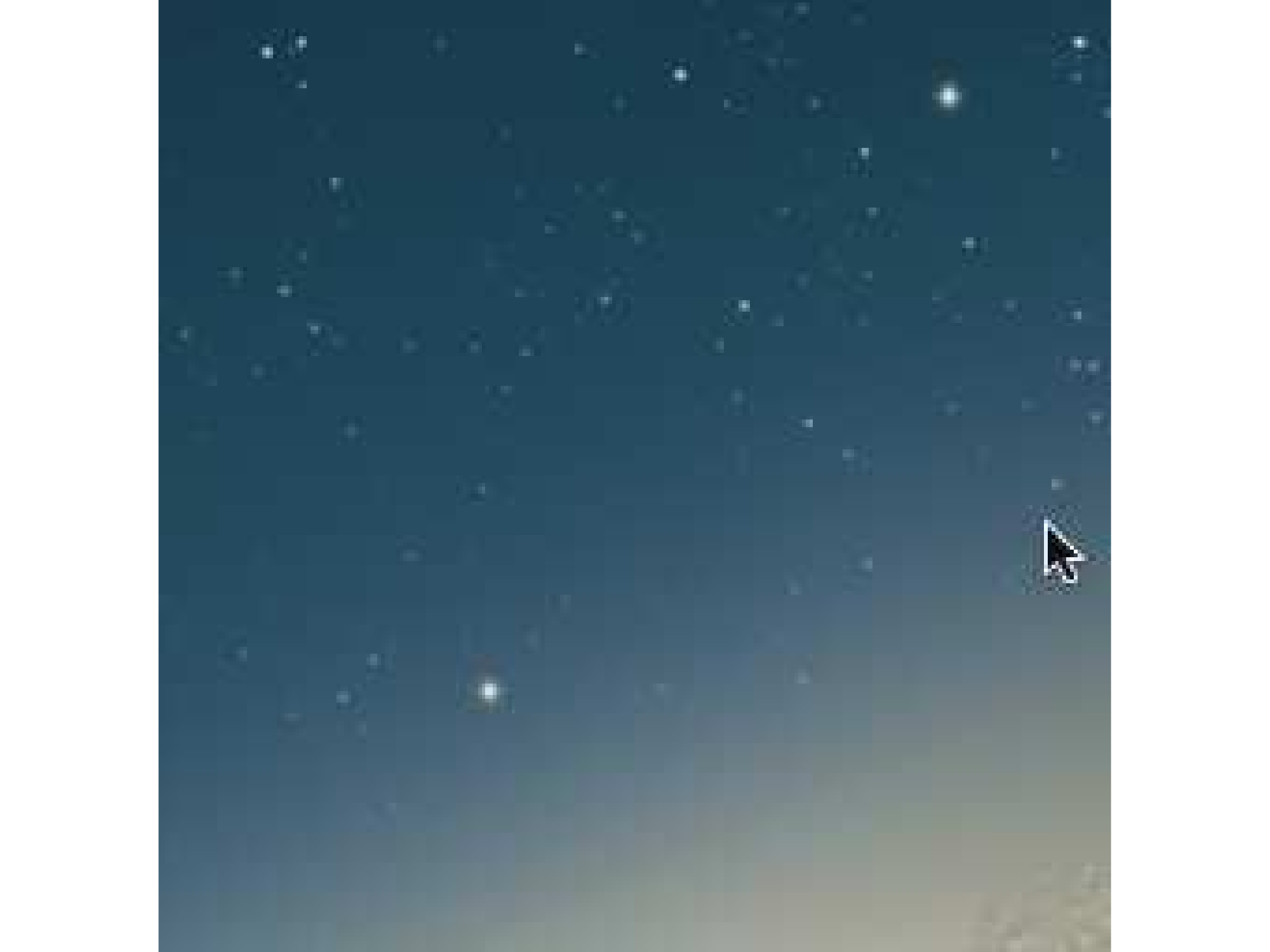}
                \vspace{-0.45cm}
            \hspace{-3cm} 
        \end{subfigure}%    
        \begin{subfigure}[b]{0.18\textwidth}
			~ %add desired spacing between images, e. g. ~, \quad, \qquad, \hfill etc.
            %(or a blank line to force the subfigure onto a new line)
                \includegraphics[width=\textwidth]{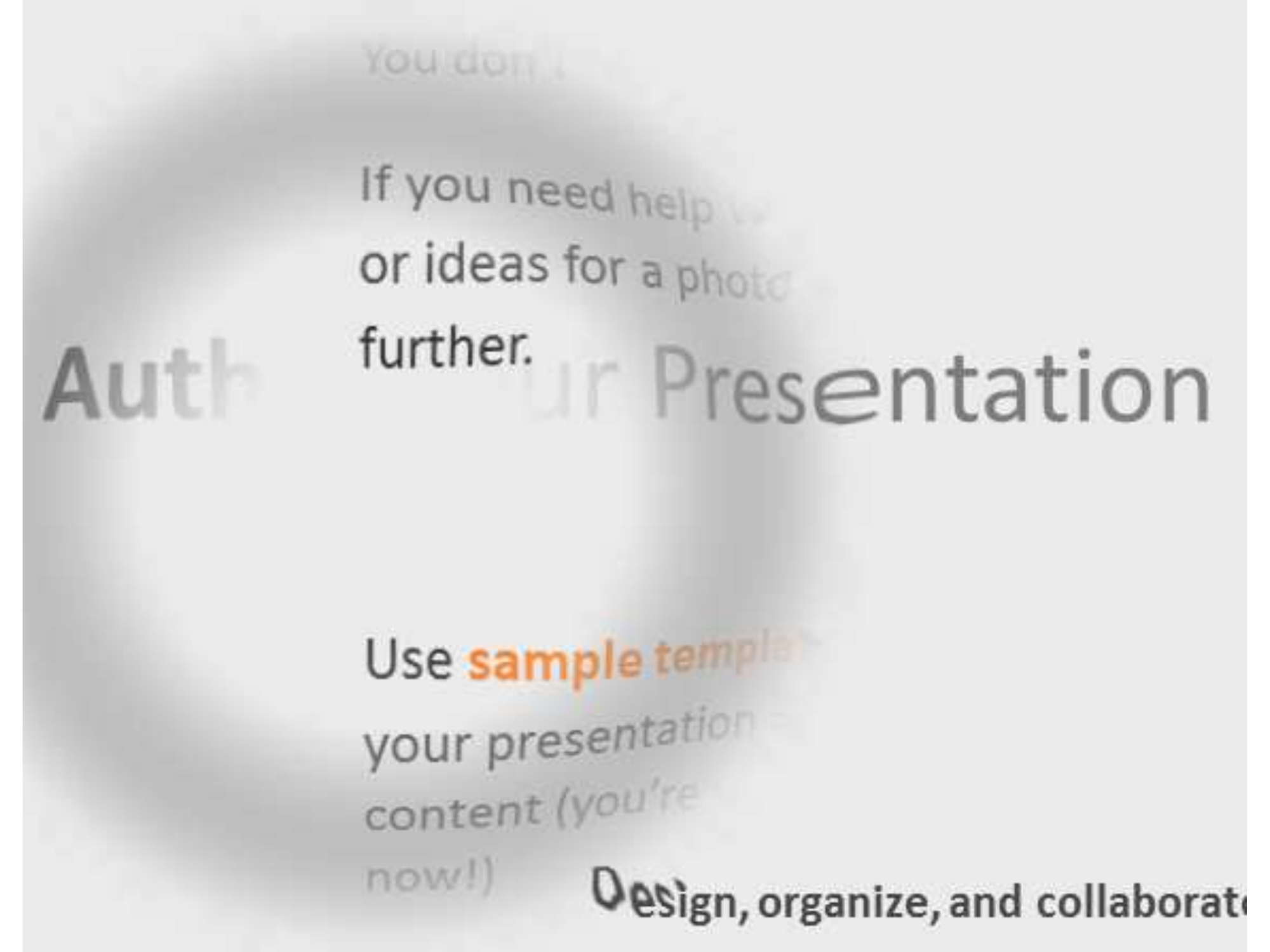}
                \vspace{-0.45cm}
            \hspace{-5cm} 
        \end{subfigure}%    
        \begin{subfigure}[b]{0.18\textwidth}
      %  \hspace{-3cm}
                \includegraphics[width=\textwidth]{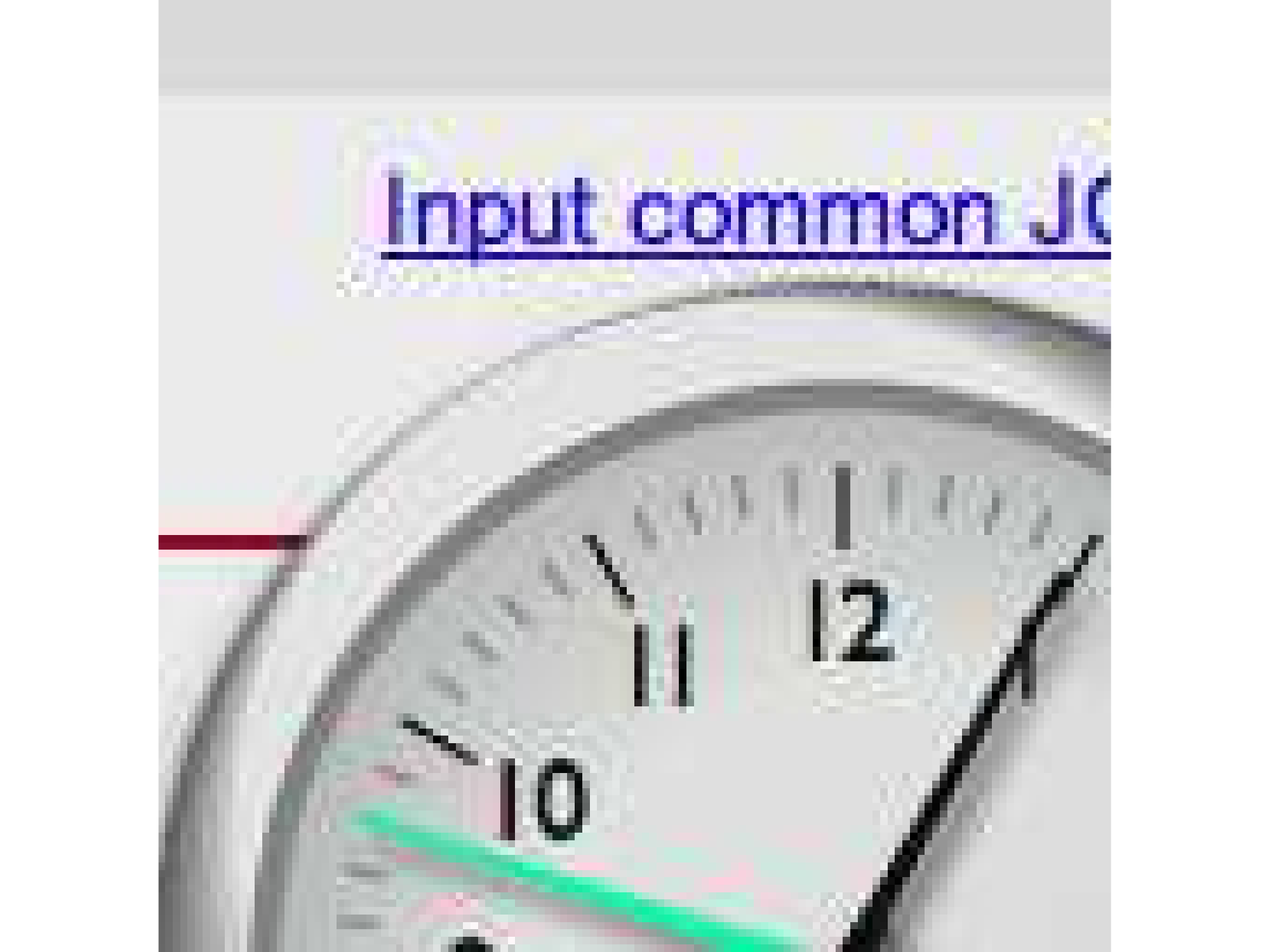}
                 \vspace{-0.45cm}
              \hspace{-4.8cm}
        \end{subfigure}
         \\[1ex]
        \begin{subfigure}[b]{0.18\textwidth}
       % \hspace{-1cm}
                \includegraphics[width=\textwidth]{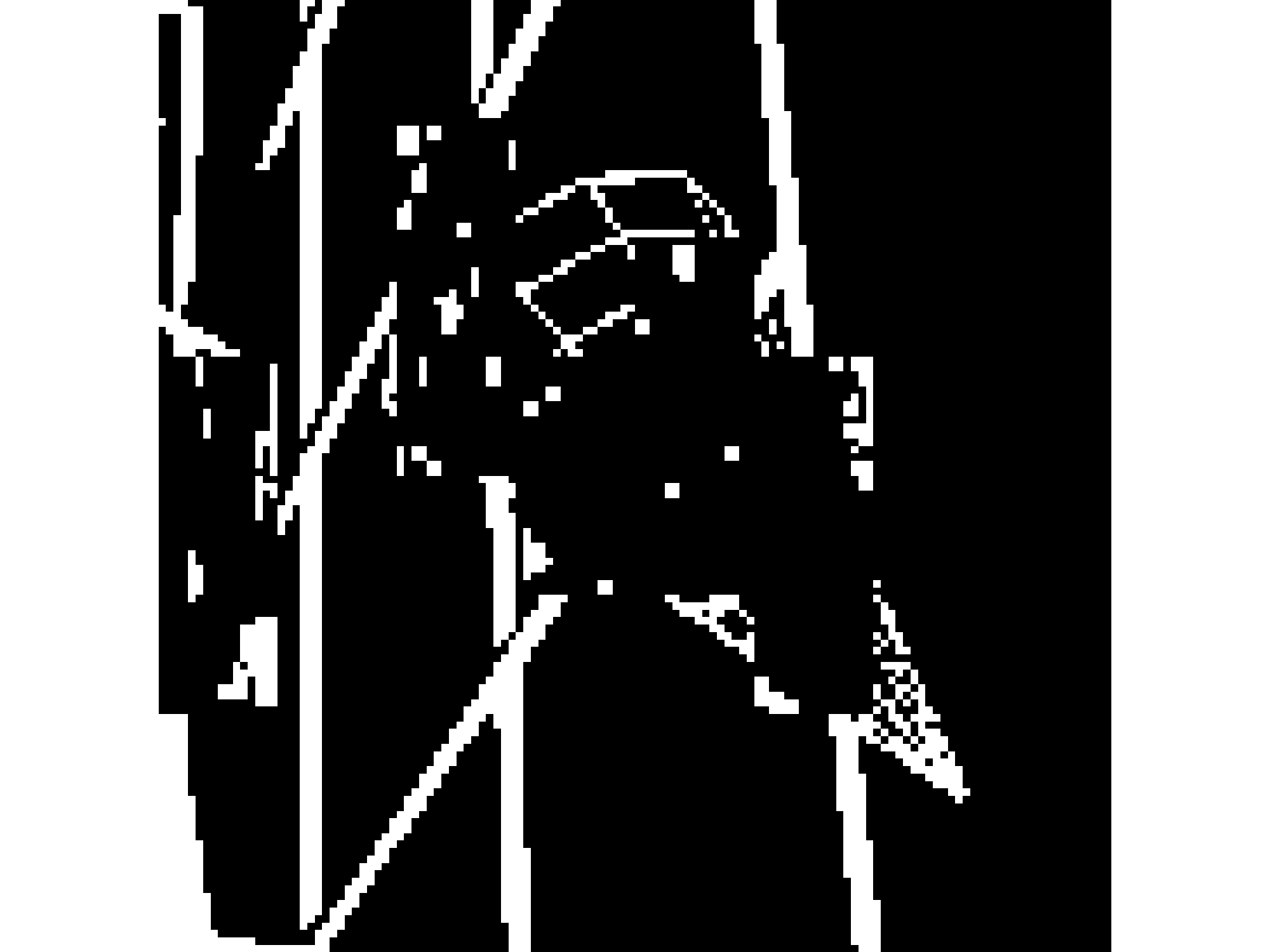}
                                \vspace{-0.5cm}
          \hspace{-2.5cm}    
        \end{subfigure}%
        ~ %add desired spacing between images, e. g. ~, \quad, \qquad, \hfill etc.
          %(or a blank line to force the subfigure onto a new line)
        \begin{subfigure}[b]{0.18\textwidth}
       % \hspace{-2cm}
                \includegraphics[width=\textwidth]{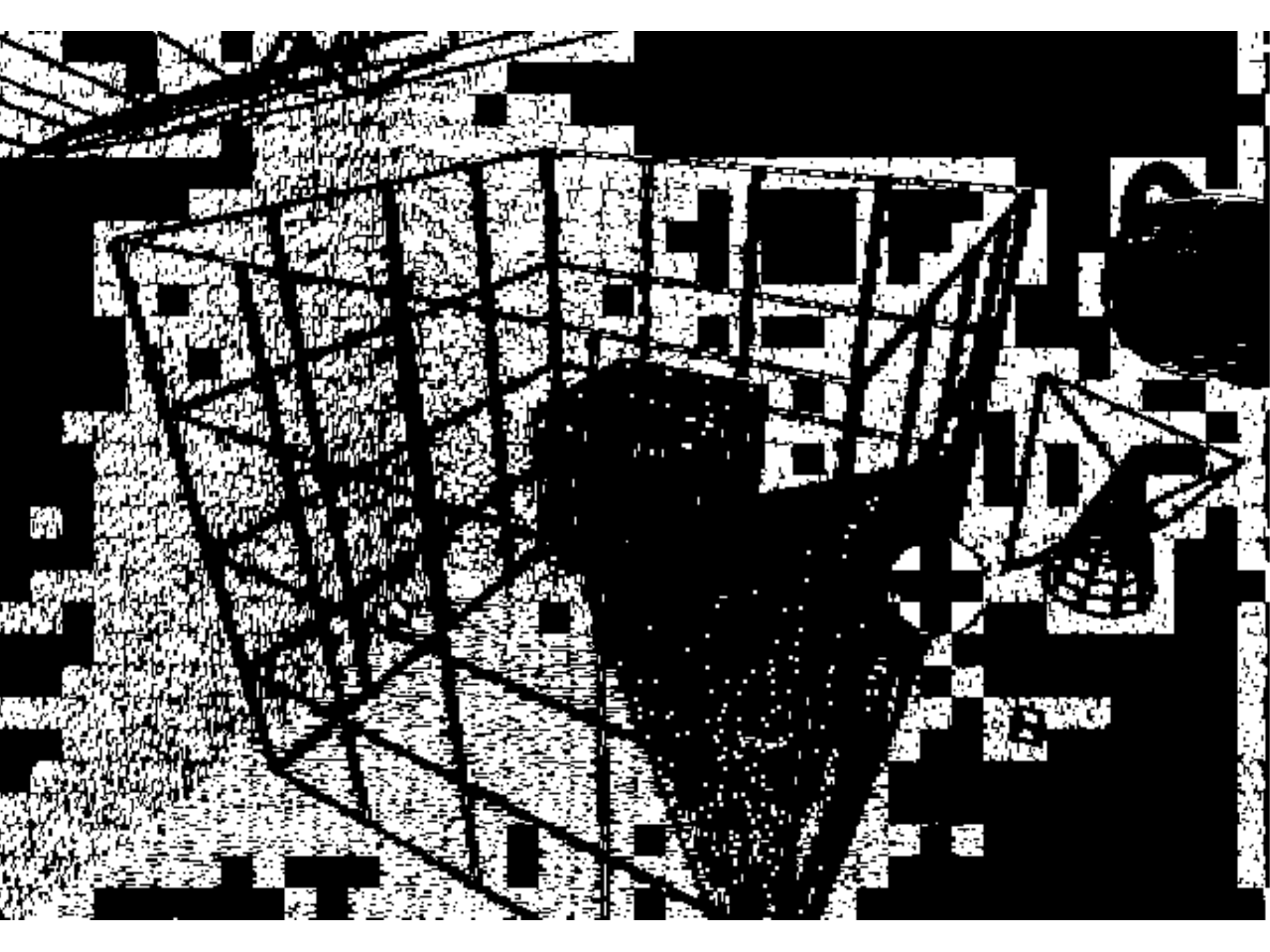}
                \vspace{-0.5cm}
            \hspace{-3cm} 
        \end{subfigure}%
        ~ %add desired spacing between images, e. g. ~, \quad, \qquad, \hfill etc.
          %(or a blank line to force the subfigure onto a new line)
        \begin{subfigure}[b]{0.18\textwidth}
       % \hspace{-2cm}
                \includegraphics[width=\textwidth]{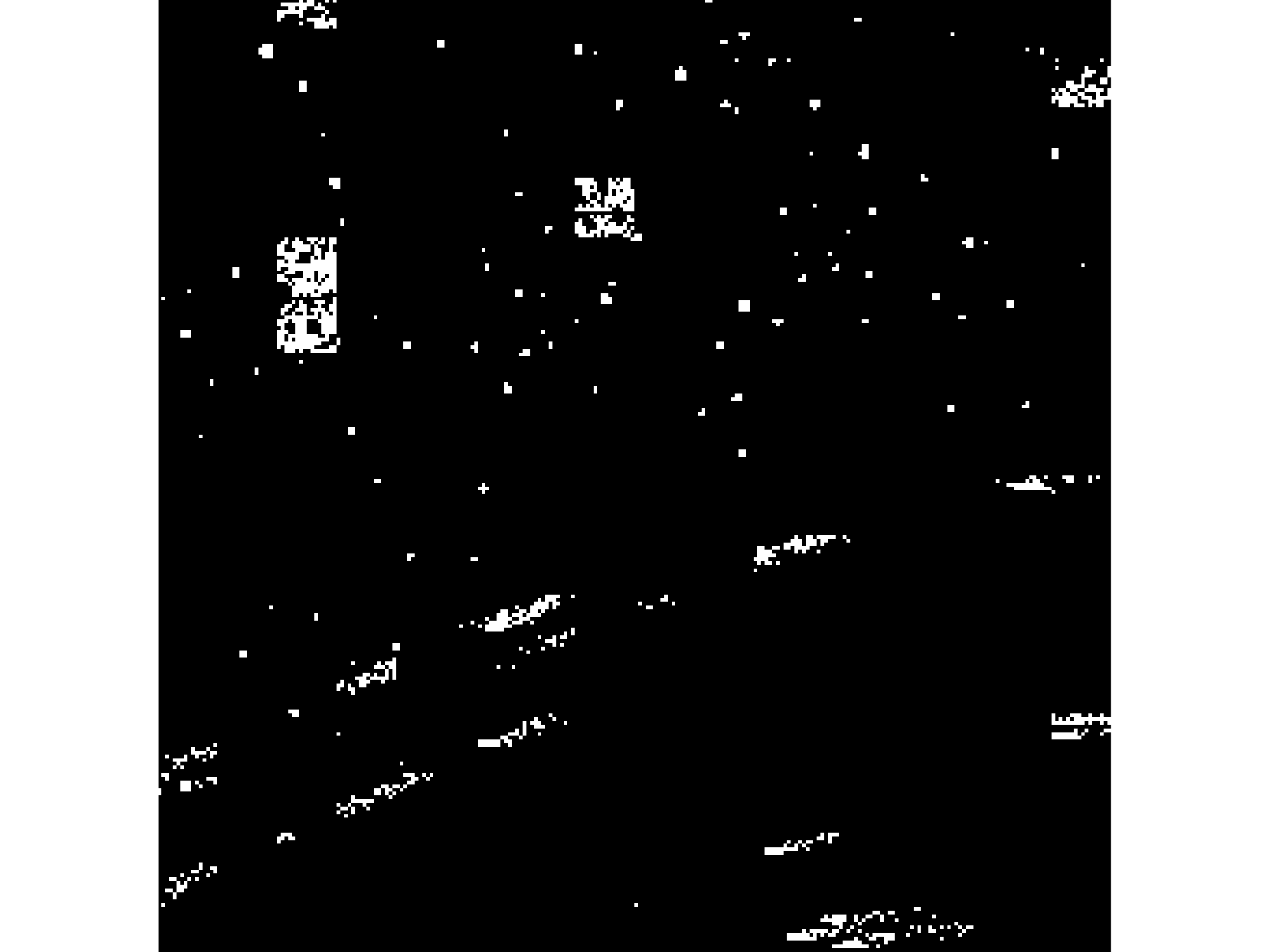}
                \vspace{-0.45cm}
            \hspace{-3cm} 
        \end{subfigure}%      
        \begin{subfigure}[b]{0.18\textwidth}
			~ %add desired spacing between images, e. g. ~, \quad, \qquad, \hfill etc.
            %(or a blank line to force the subfigure onto a new line)
                \includegraphics[width=\textwidth]{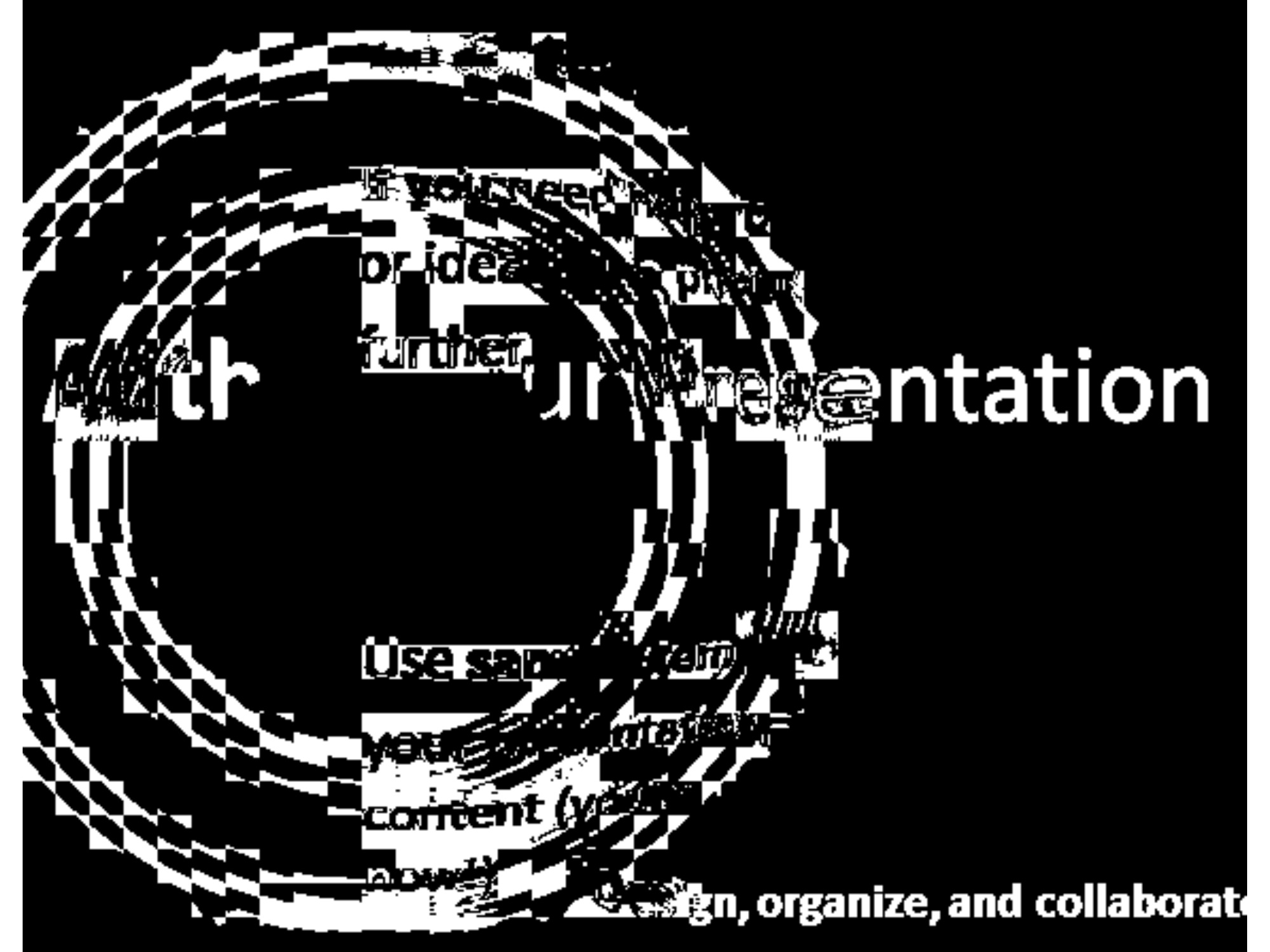}
                \vspace{-0.45cm}
            \hspace{-3cm} 
        \end{subfigure}%            
        \begin{subfigure}[b]{0.18\textwidth}
      %  \hspace{-3cm}
                \includegraphics[width=\textwidth]{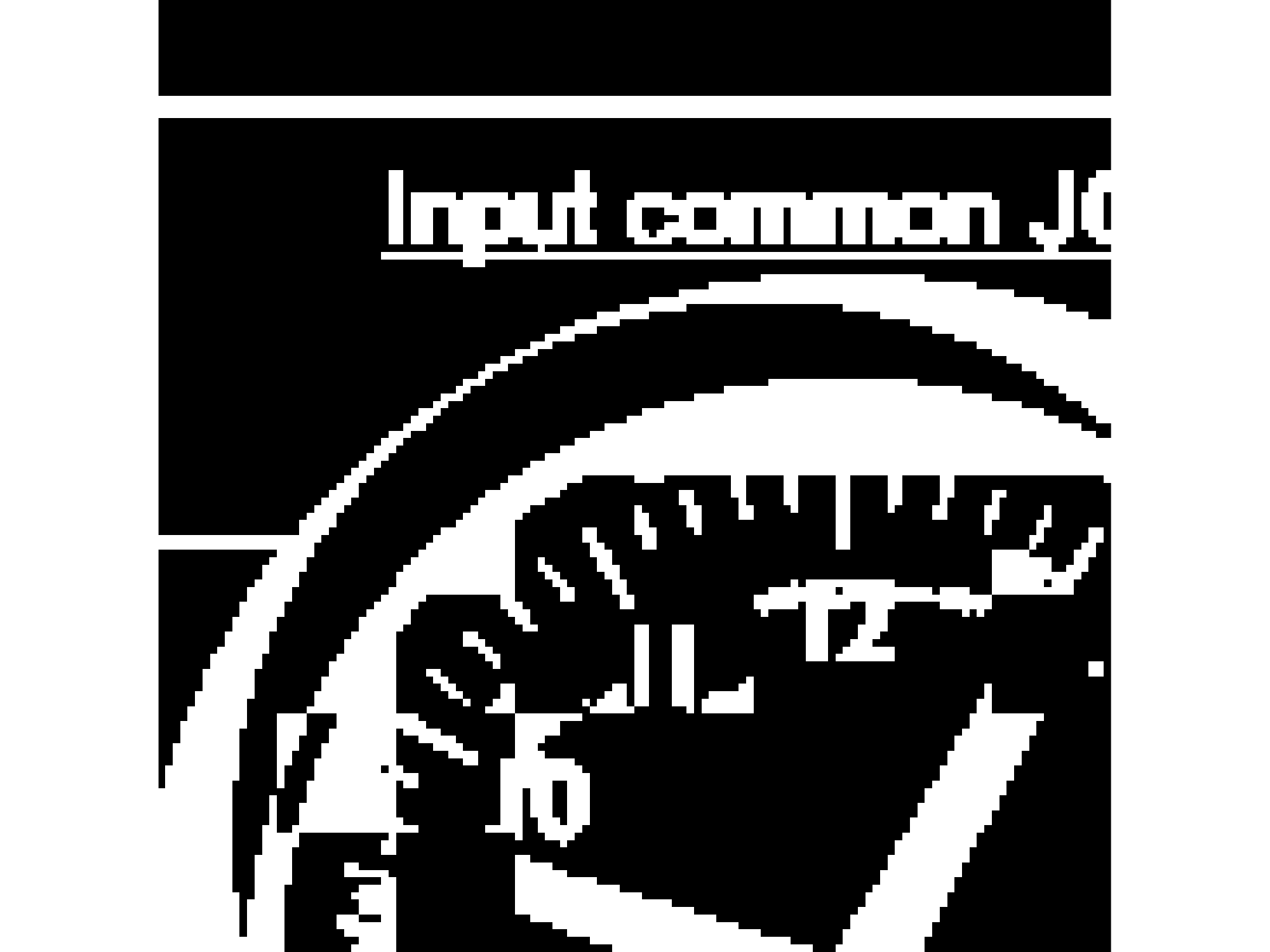}
                 \vspace{-0.45cm}
              \hspace{-4.8cm}
        \end{subfigure} \\[1ex]
        \begin{subfigure}[b]{0.18\textwidth}
       % \hspace{-1cm}
                \includegraphics[width=\textwidth]{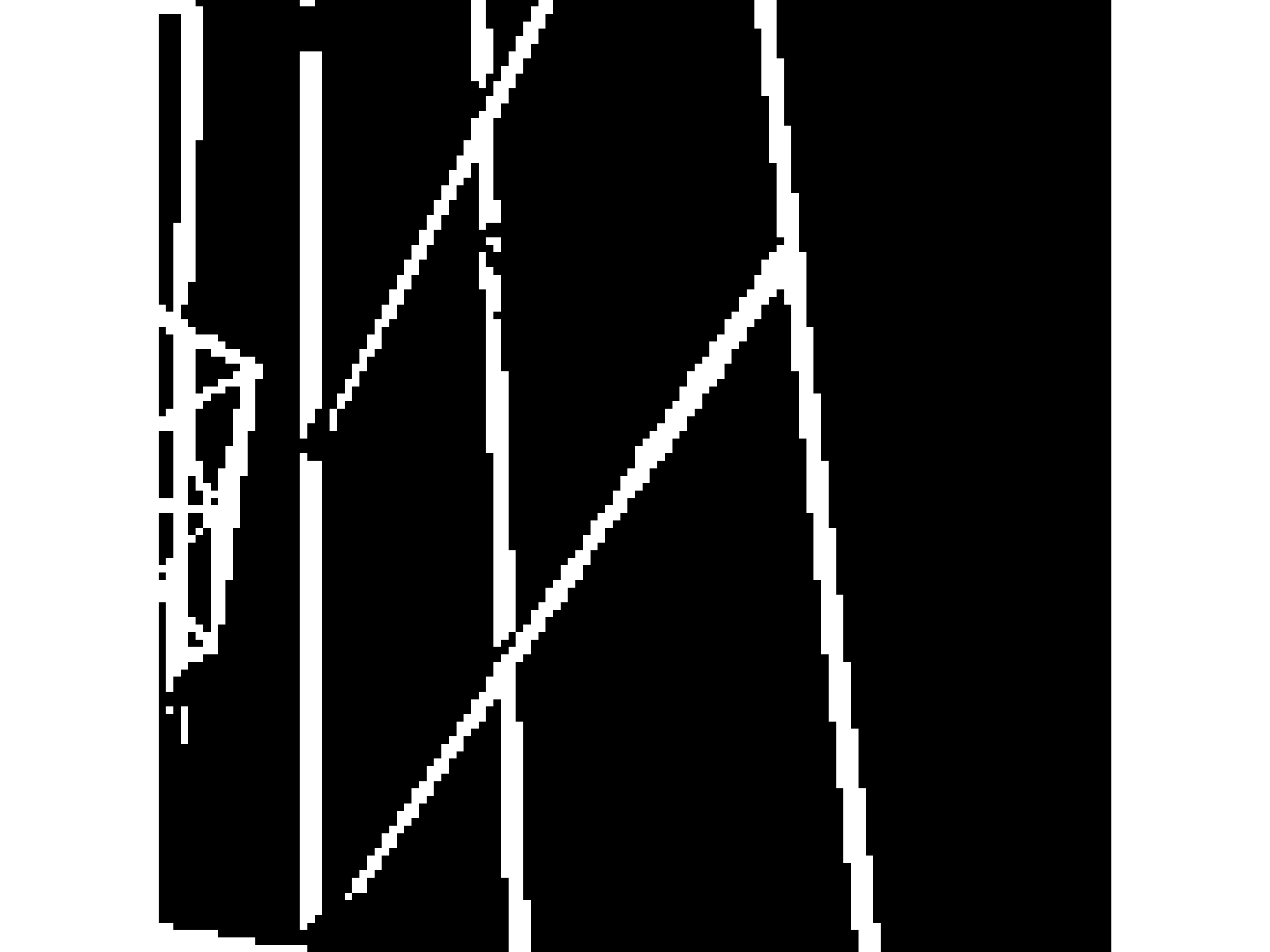}
                                \vspace{-0.5cm}
          \hspace{-2.5cm}    
        \end{subfigure}%
        ~ %add desired spacing between images, e. g. ~, \quad, \qquad, \hfill etc.
          %(or a blank line to force the subfigure onto a new line)
        \begin{subfigure}[b]{0.18\textwidth}
       % \hspace{-2cm}
                \includegraphics[width=\textwidth]{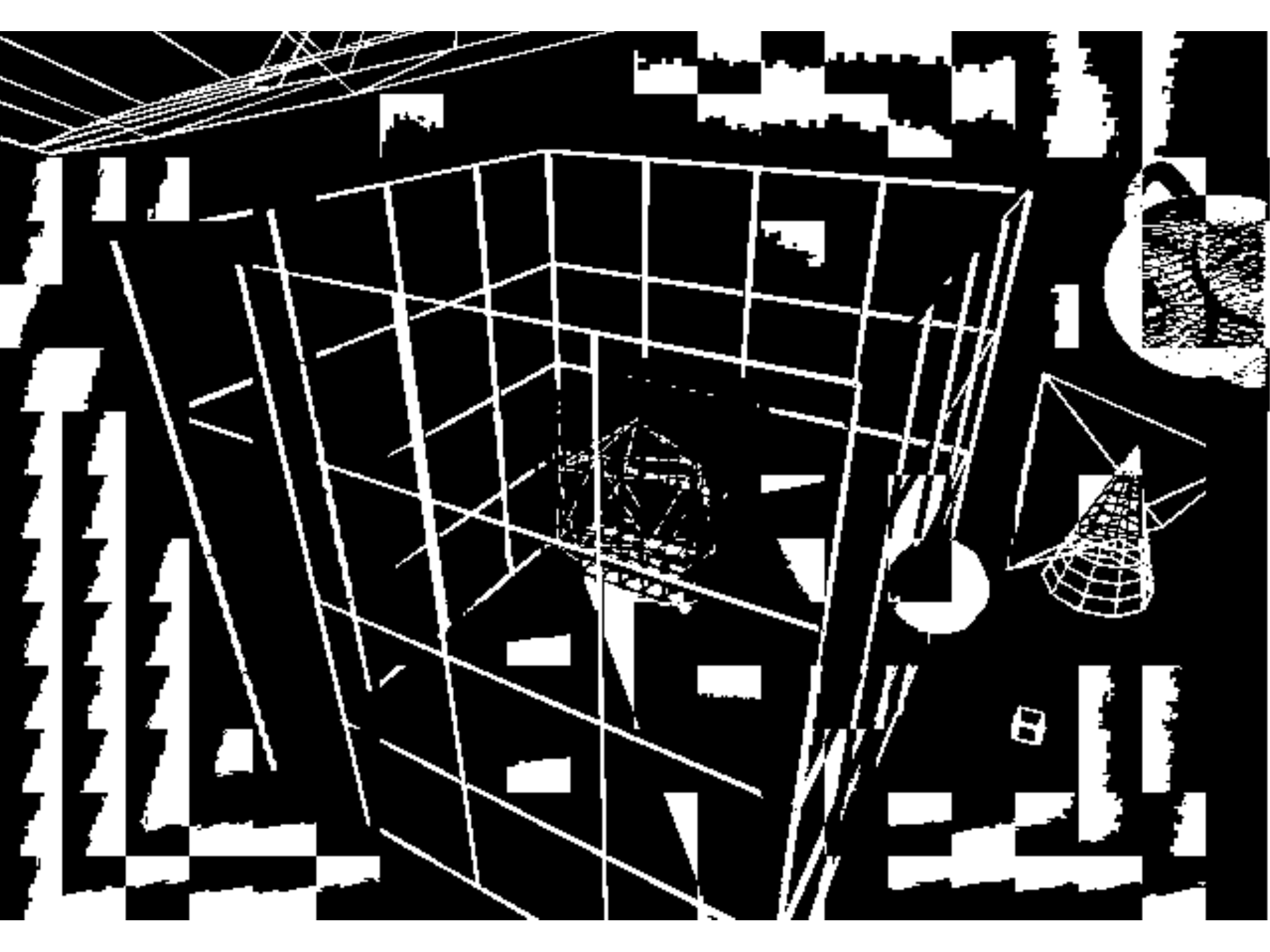}
                \vspace{-0.5cm}
            \hspace{-3cm} 
        \end{subfigure}%
        ~ %add desired spacing between images, e. g. ~, \quad, \qquad, \hfill etc.
          %(or a blank line to force the subfigure onto a new line)
        \begin{subfigure}[b]{0.18\textwidth}
       % \hspace{-2cm}
                \includegraphics[width=\textwidth]{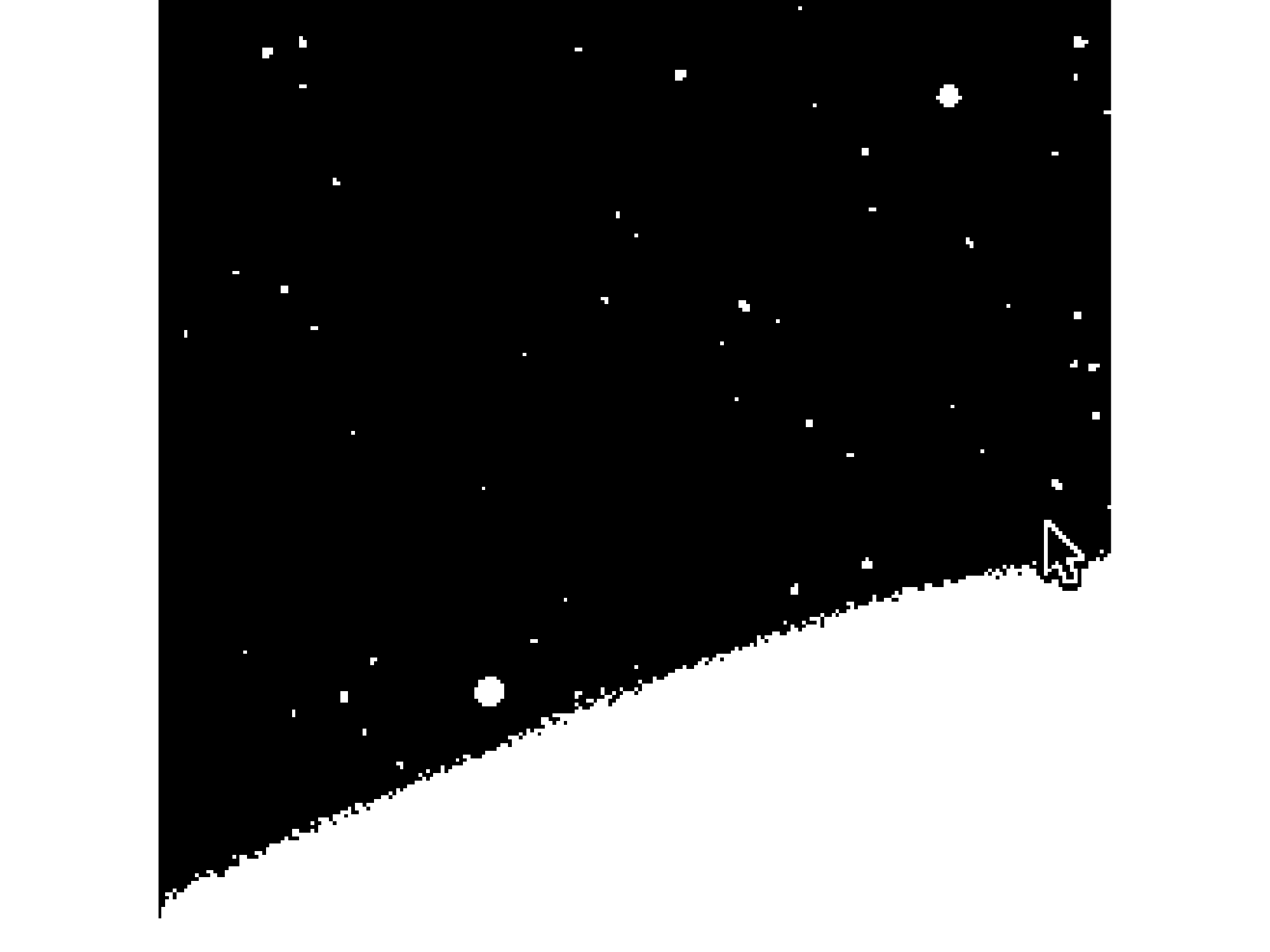}
                \vspace{-0.45cm}
            \hspace{-3cm} 
        \end{subfigure}%   
        \begin{subfigure}[b]{0.18\textwidth}
			~ %add desired spacing between images, e. g. ~, \quad, \qquad, \hfill etc.
            %(or a blank line to force the subfigure onto a new line)
                \includegraphics[width=\textwidth]{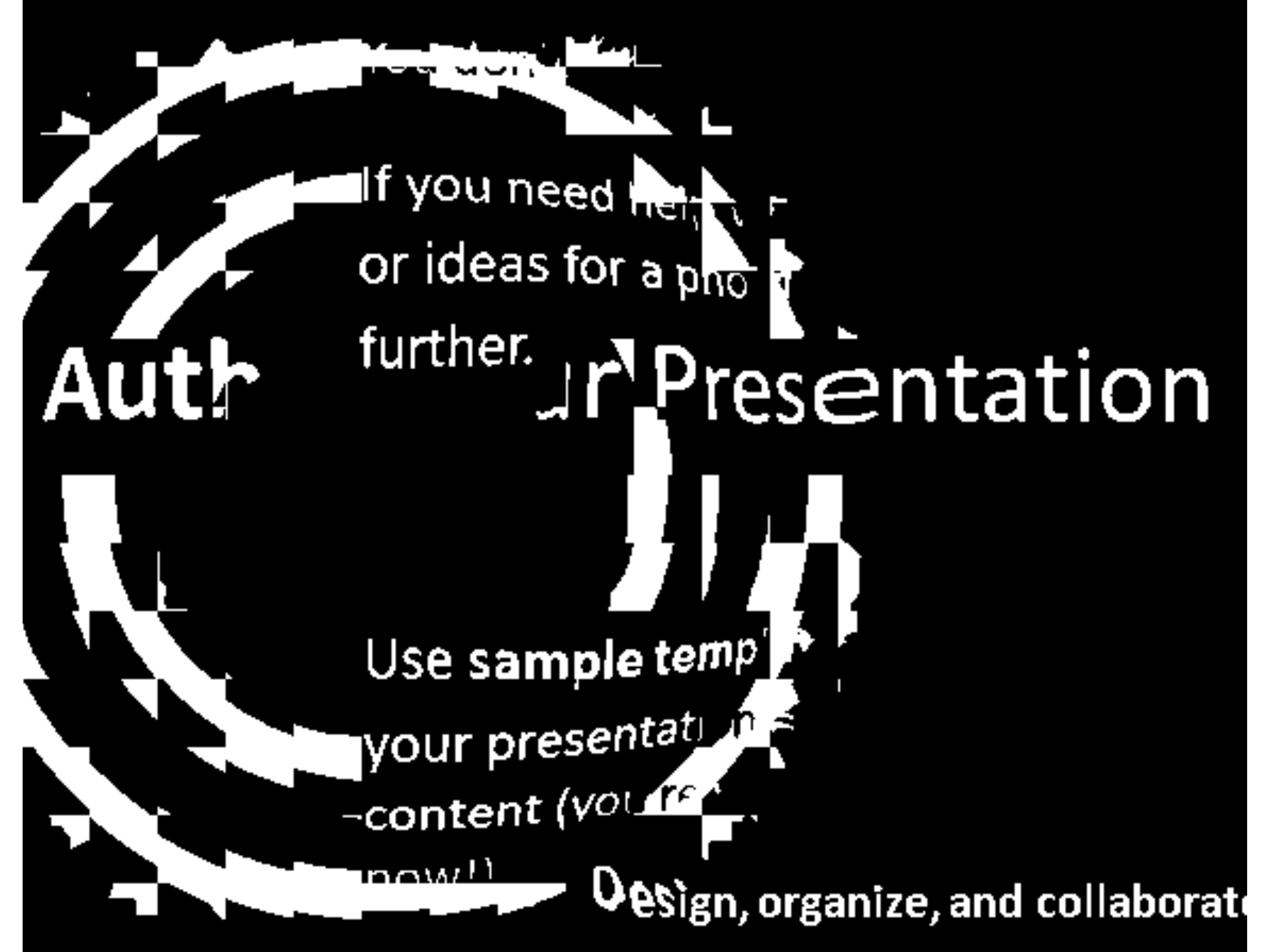}
                \vspace{-0.45cm}
            \hspace{-3cm} 
        \end{subfigure}%            
        \begin{subfigure}[b]{0.18\textwidth}
      %  \hspace{-3cm}
                \includegraphics[width=\textwidth]{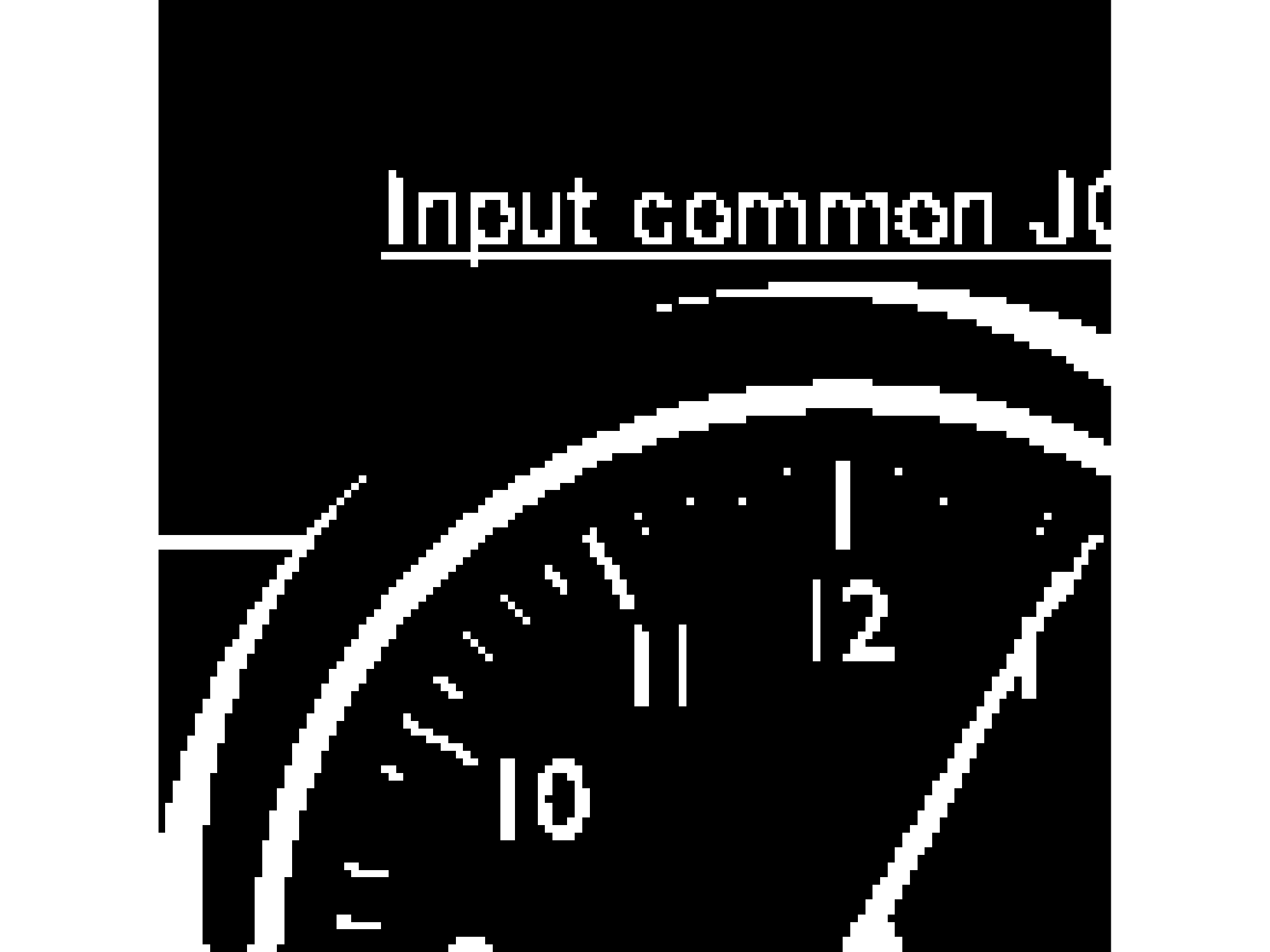}
                 \vspace{-0.45cm}
              \hspace{-4.8cm}
        \end{subfigure}\\[1ex]        
        \begin{subfigure}[b]{0.18\textwidth}
       % \hspace{-1cm}
                \includegraphics[width=\textwidth]{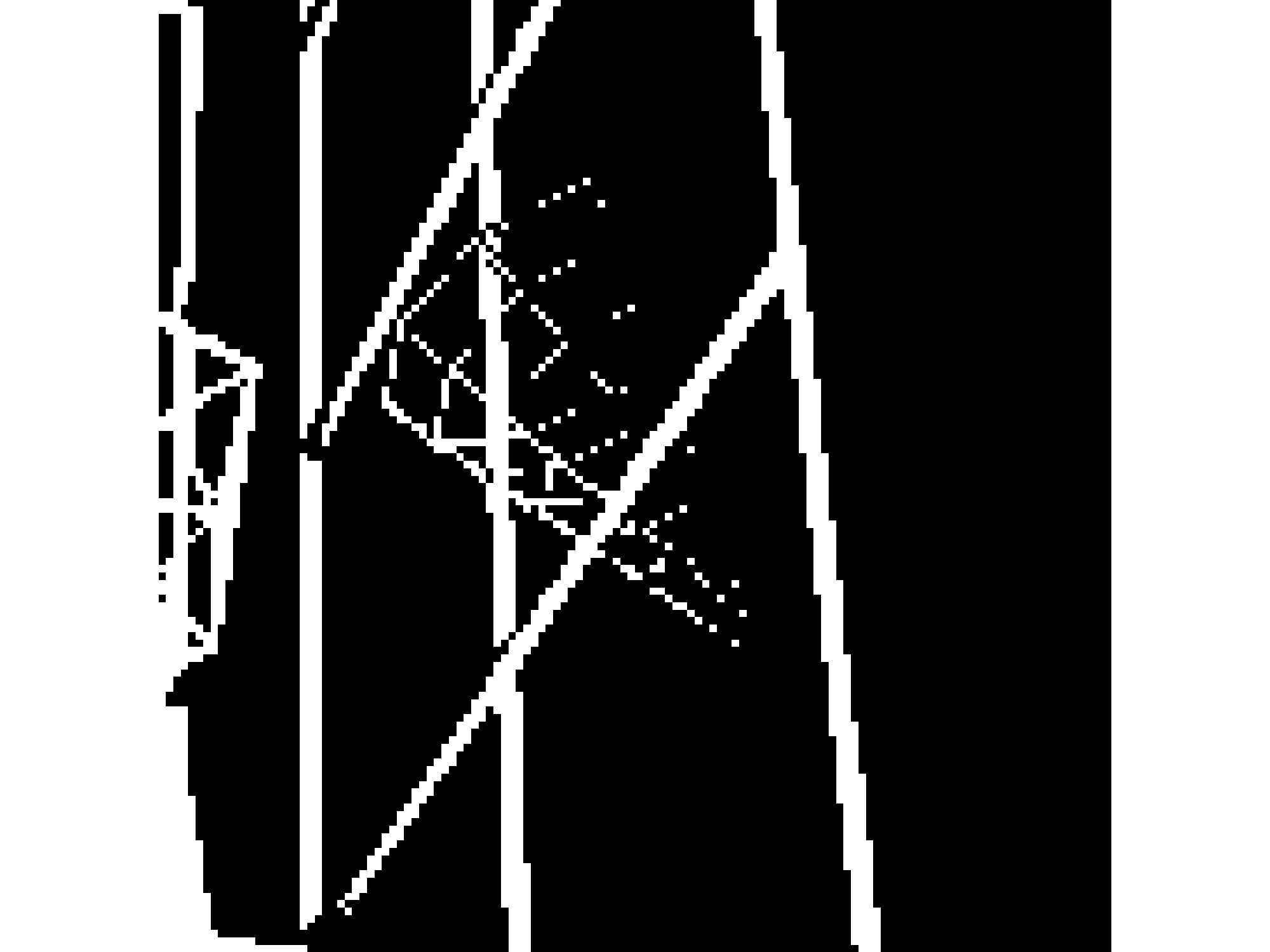}
                                \vspace{-0.5cm}
          \hspace{-2.5cm}    
        \end{subfigure}%
        ~ %add desired spacing between images, e. g. ~, \quad, \qquad, \hfill etc.
          %(or a blank line to force the subfigure onto a new line)
        \begin{subfigure}[b]{0.18\textwidth}
       % \hspace{-2cm}
                \includegraphics[width=\textwidth]{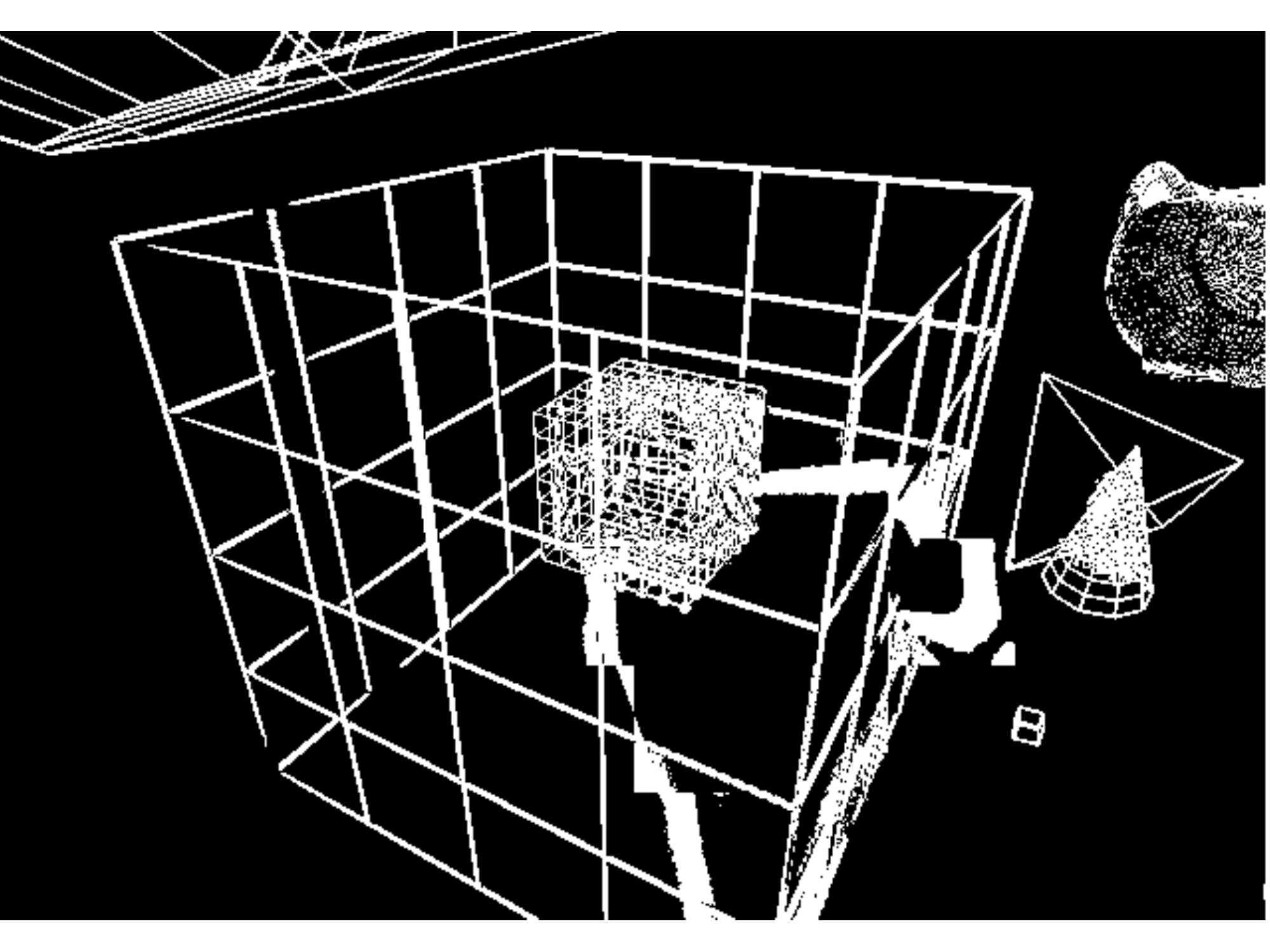}
                \vspace{-0.5cm}
            \hspace{-3cm} 
        \end{subfigure}%
        ~ %add desired spacing between images, e. g. ~, \quad, \qquad, \hfill etc.
          %(or a blank line to force the subfigure onto a new line)
        \begin{subfigure}[b]{0.18\textwidth}
       % \hspace{-2cm}
                \includegraphics[width=\textwidth]{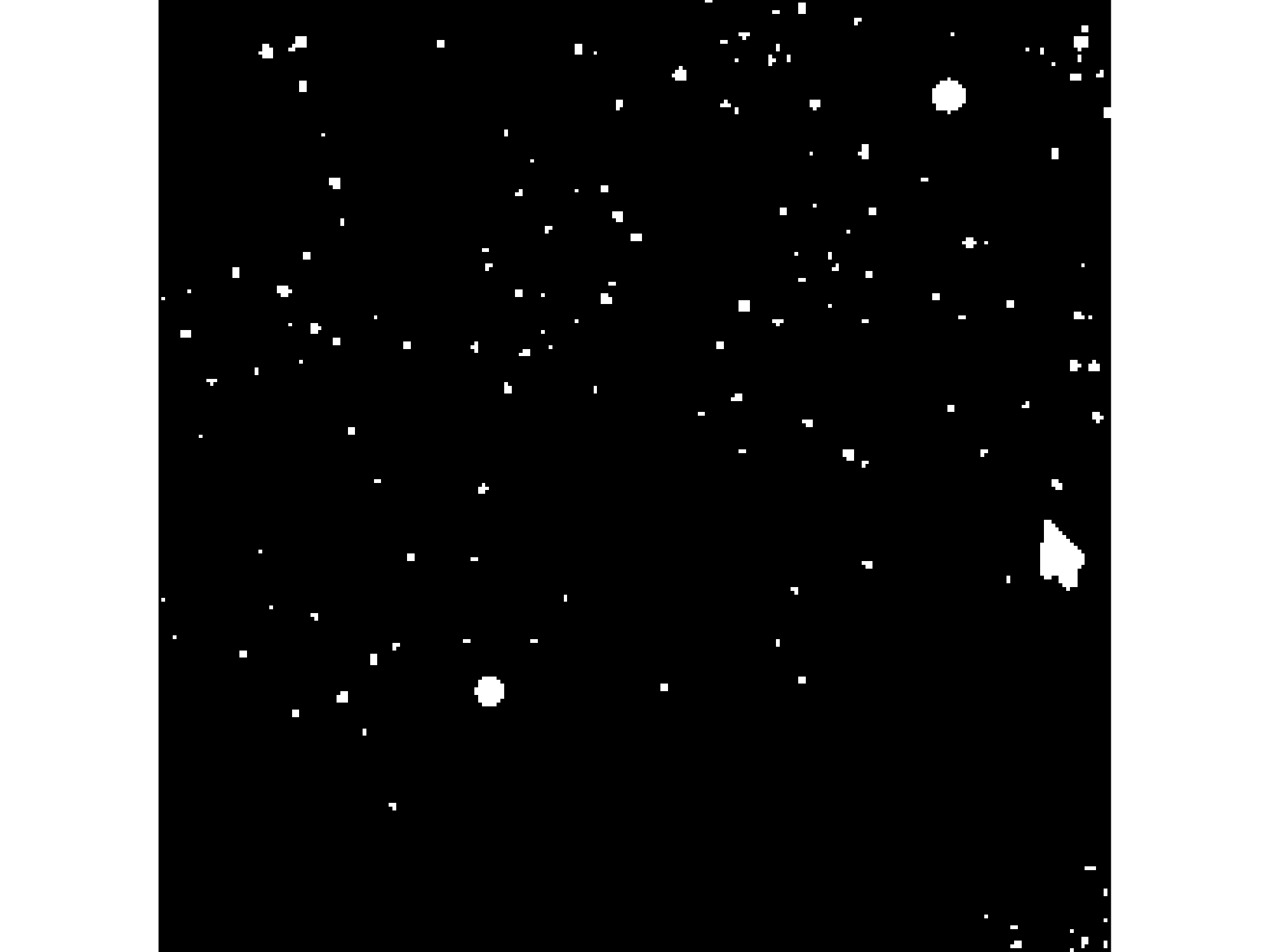}
                \vspace{-0.45cm}
            \hspace{-3cm} 
        \end{subfigure}%
        \begin{subfigure}[b]{0.18\textwidth}
			~ %add desired spacing between images, e. g. ~, \quad, \qquad, \hfill etc.
            %(or a blank line to force the subfigure onto a new line)
                \includegraphics[width=\textwidth]{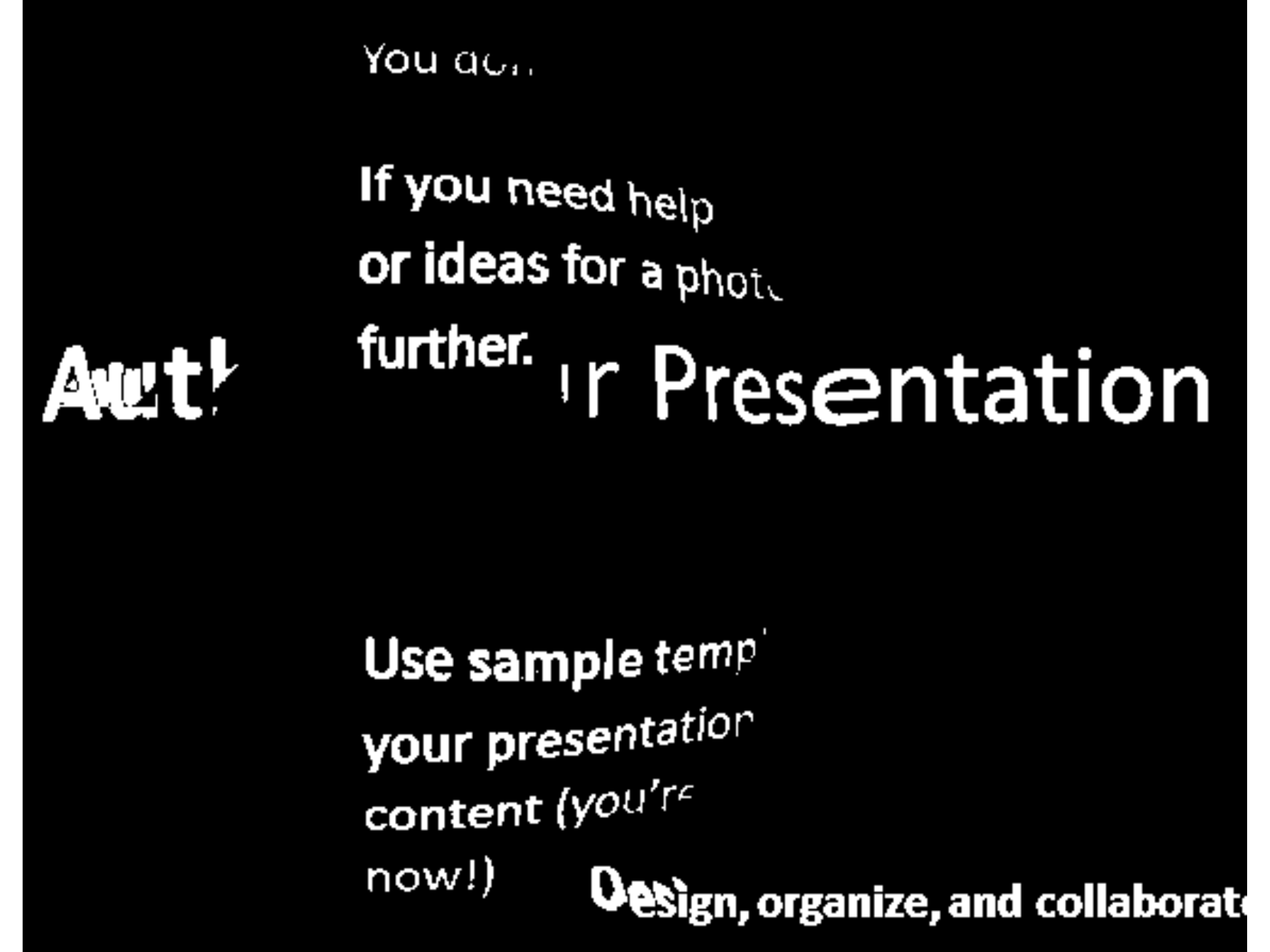}
                \vspace{-0.45cm}
            \hspace{-3cm} 
        \end{subfigure}%                  
        \begin{subfigure}[b]{0.18\textwidth}
      %  \hspace{-3cm}
                \includegraphics[width=\textwidth]{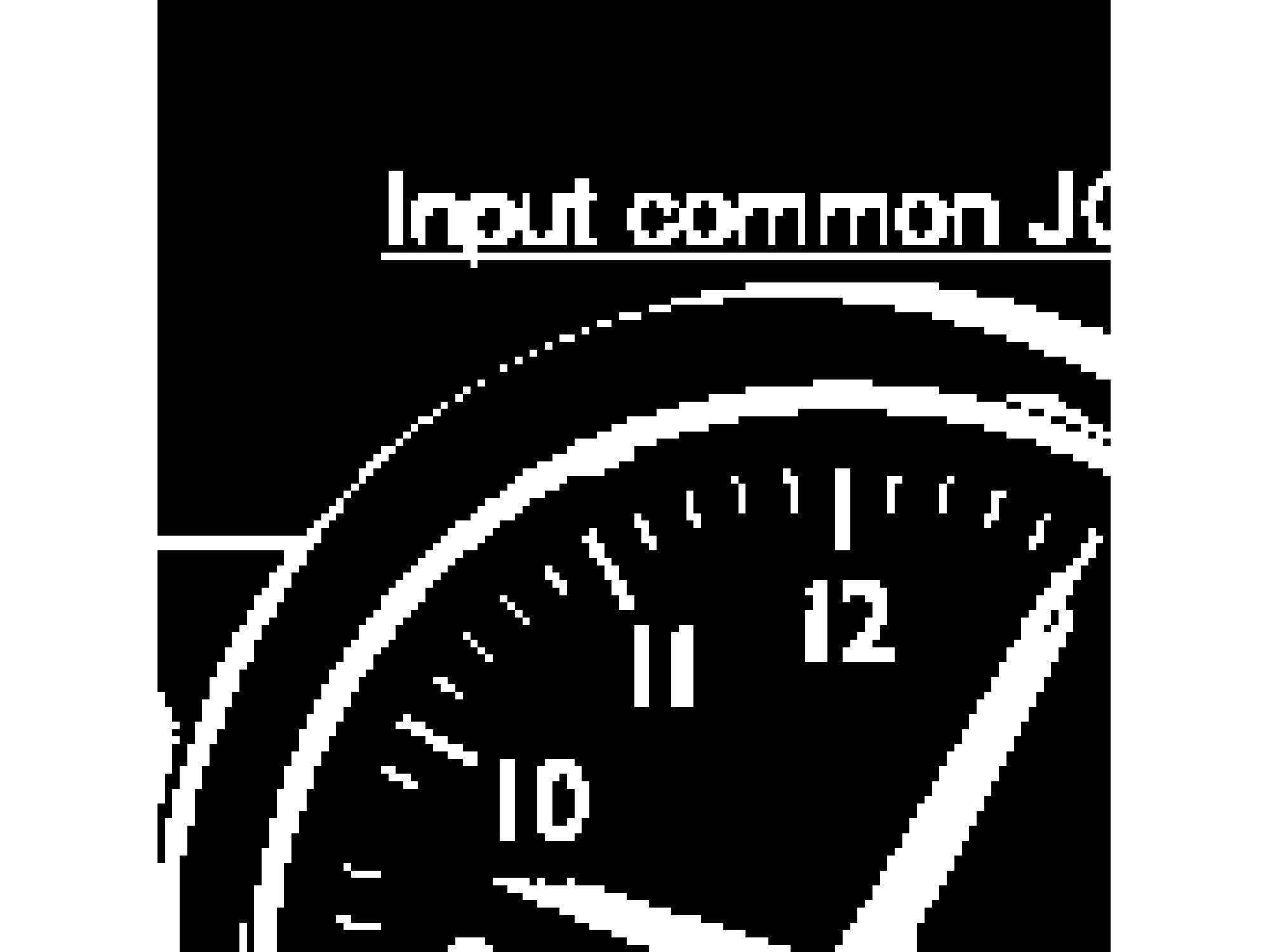}
                 \vspace{-0.45cm}
              \hspace{-4.8cm}
        \end{subfigure} \\[1ex]      
        \begin{subfigure}[b]{0.18\textwidth}
       % \hspace{-1cm}
                \includegraphics[width=\textwidth]{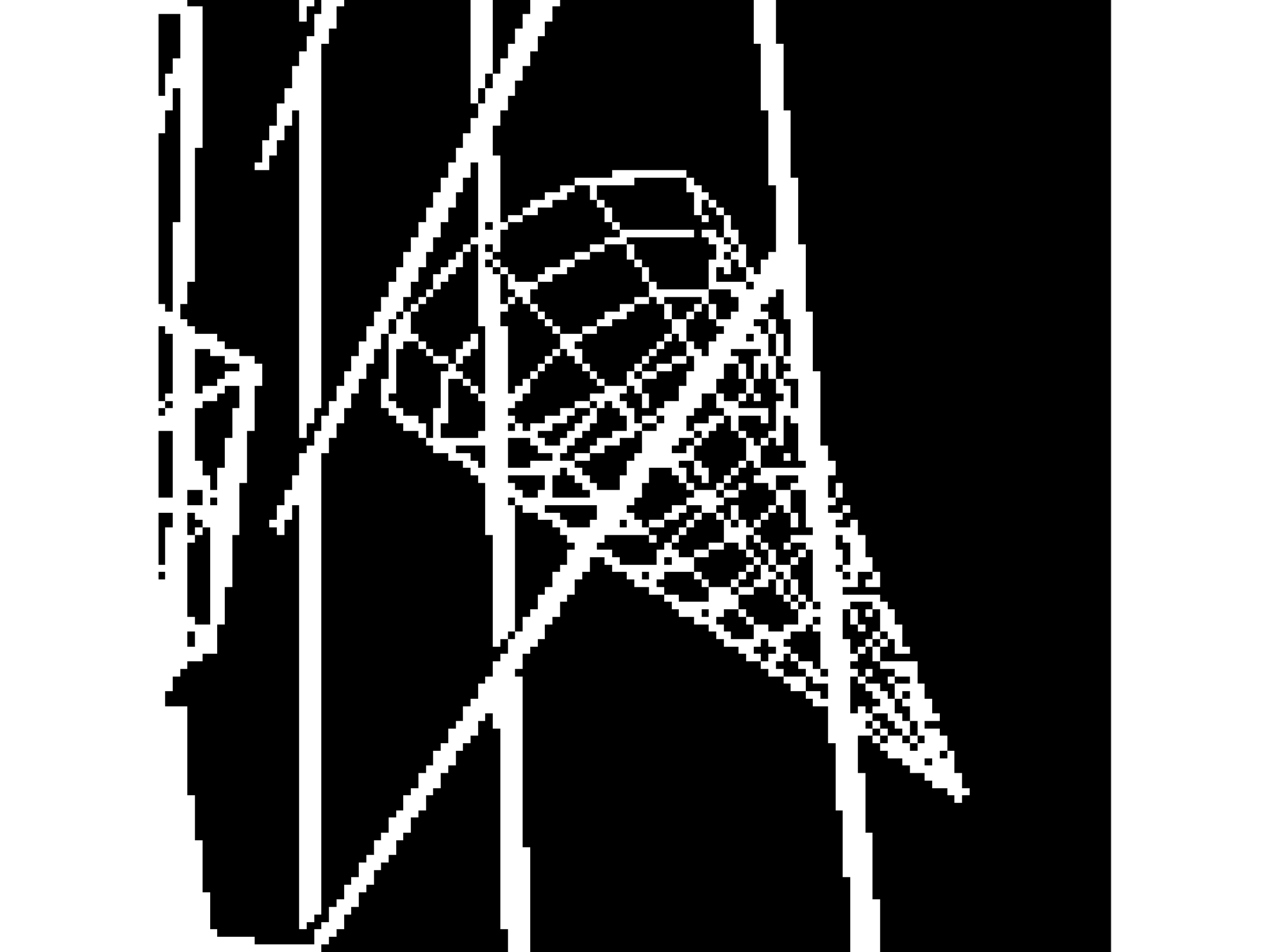}
                                \vspace{-0.5cm}
          \hspace{-2.5cm}    
        \end{subfigure}%
        ~ %add desired spacing between images, e. g. ~, \quad, \qquad, \hfill etc.
          %(or a blank line to force the subfigure onto a new line)
        \begin{subfigure}[b]{0.18\textwidth}
       % \hspace{-2cm}
                \includegraphics[width=\textwidth]{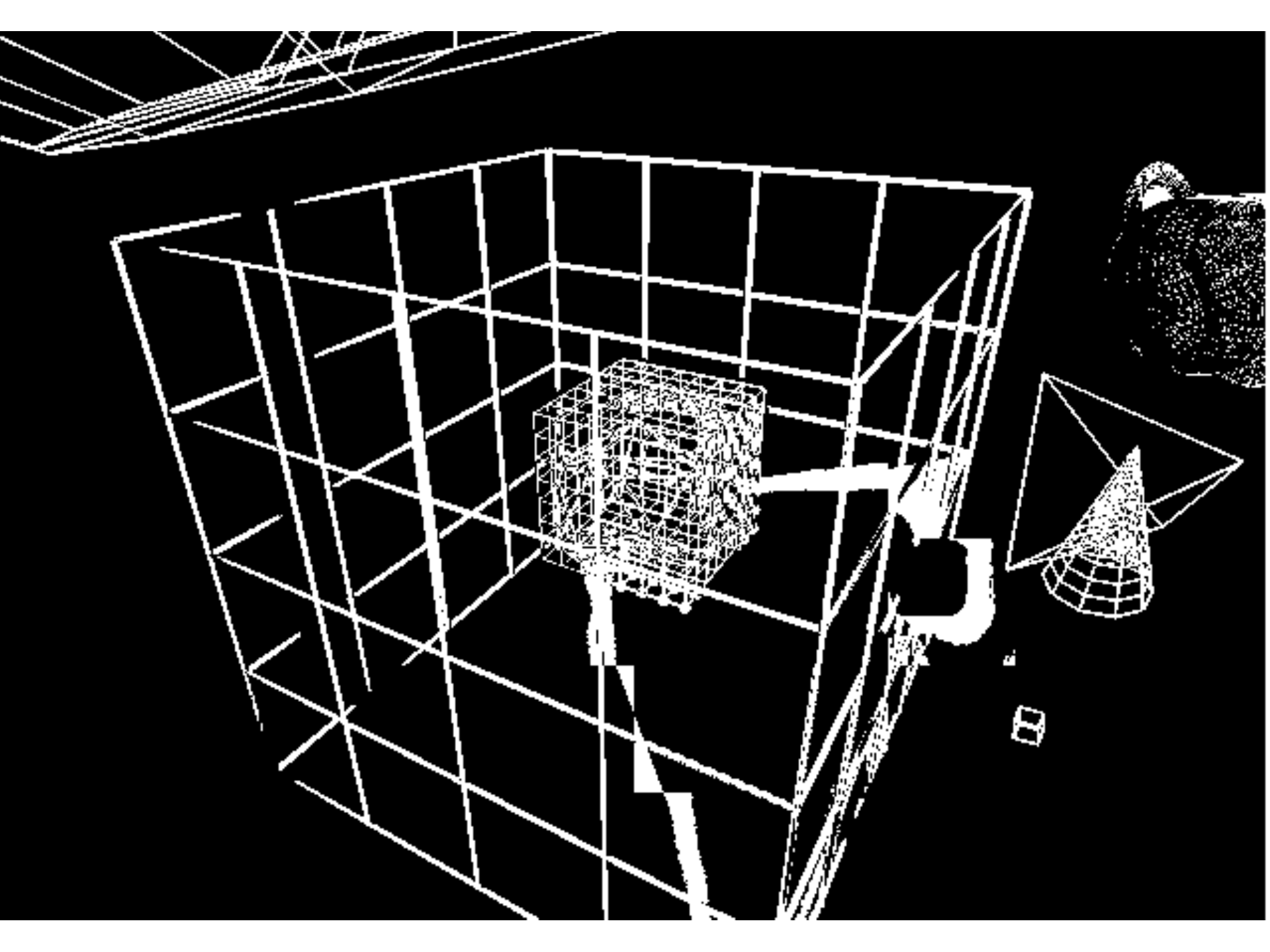}
                \vspace{-0.5cm}
            \hspace{-3cm} 
        \end{subfigure}%
        ~ %add desired spacing between images, e. g. ~, \quad, \qquad, \hfill etc.
          %(or a blank line to force the subfigure onto a new line)
        \begin{subfigure}[b]{0.18\textwidth}
       % \hspace{-2cm}
                \includegraphics[width=\textwidth]{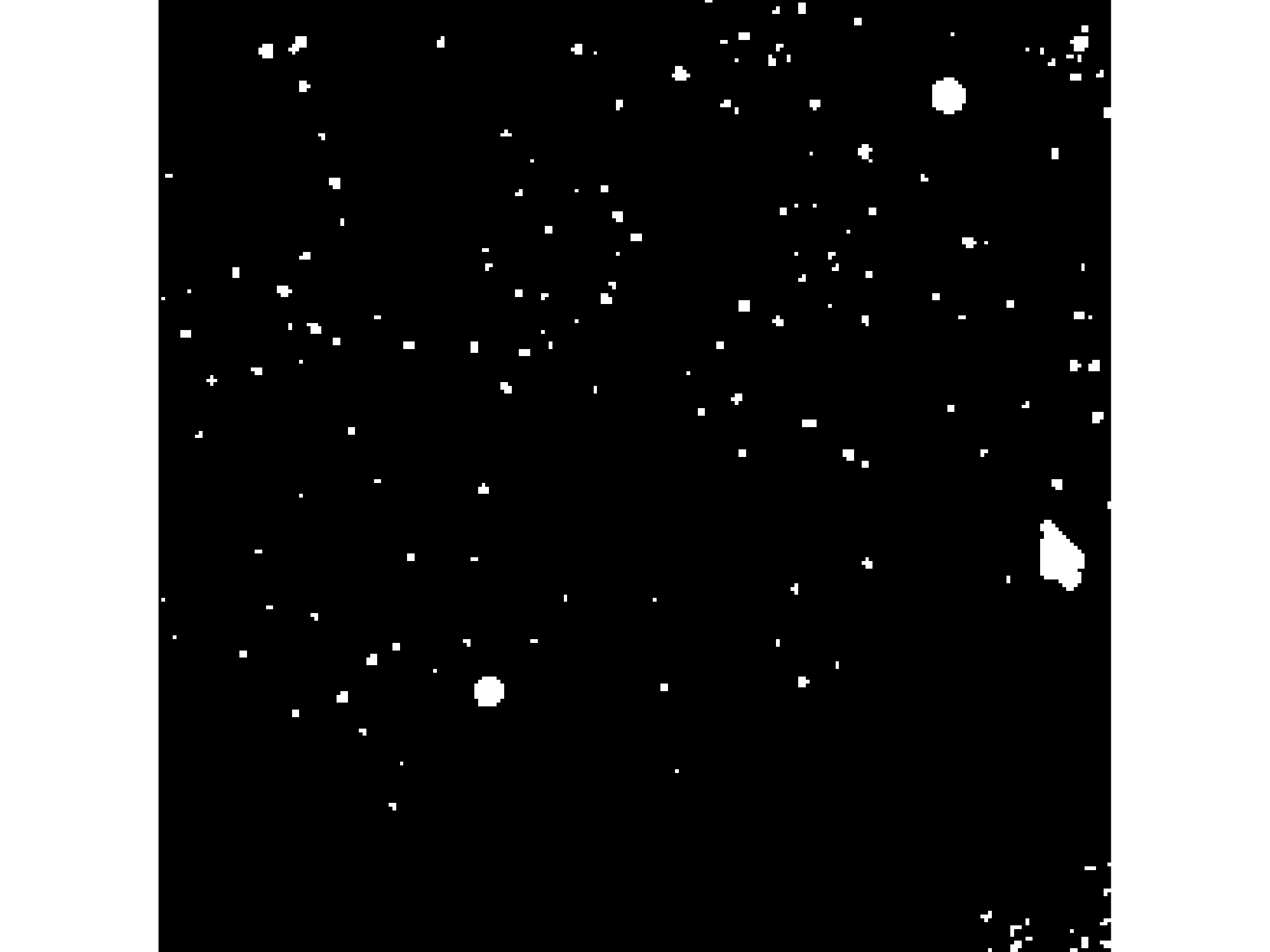}
                \vspace{-0.45cm}
            \hspace{-3cm} 
        \end{subfigure}%
        \begin{subfigure}[b]{0.18\textwidth}
			~ %add desired spacing between images, e. g. ~, \quad, \qquad, \hfill etc.
            %(or a blank line to force the subfigure onto a new line)
                \includegraphics[width=\textwidth]{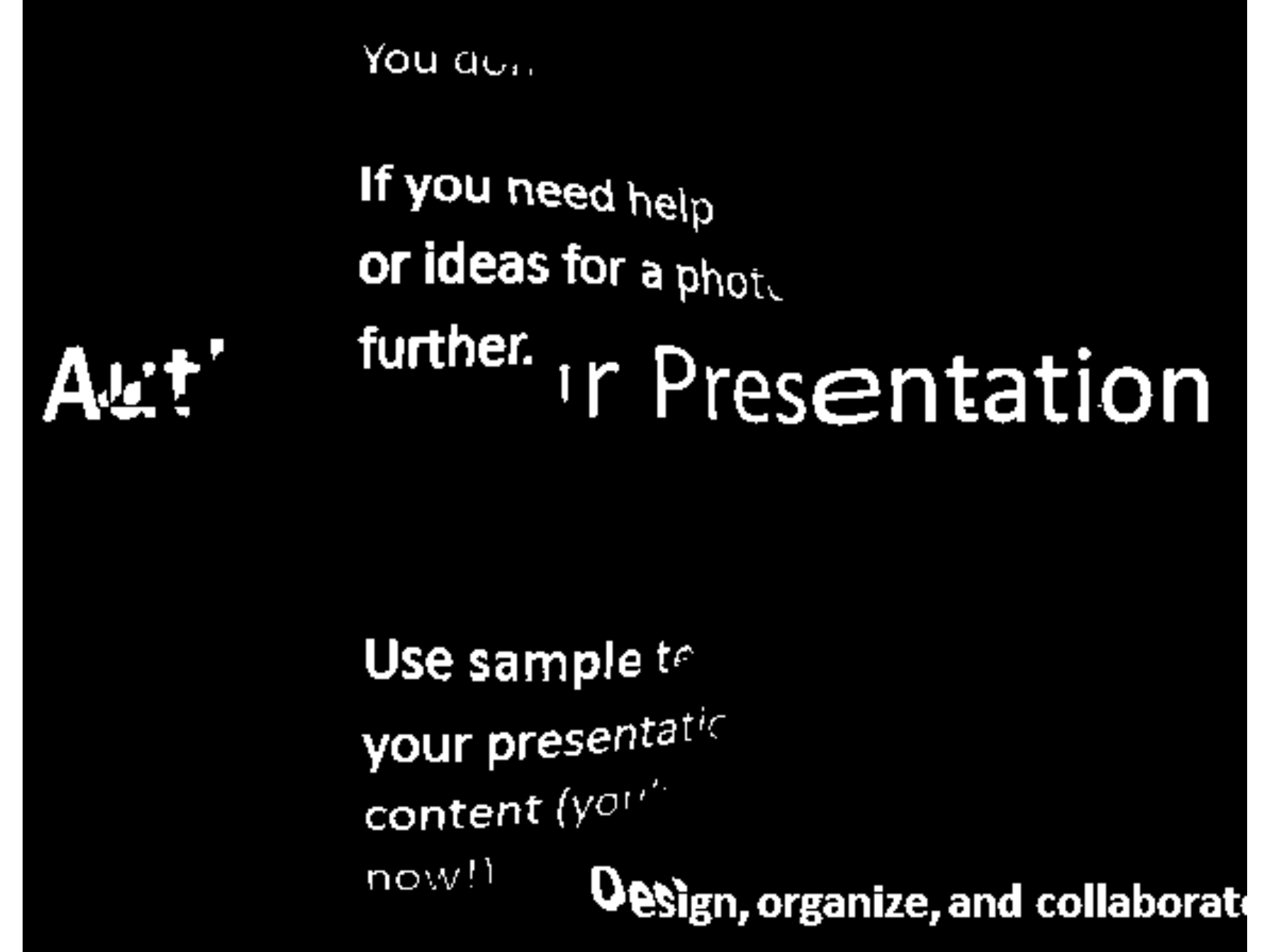} %% 2_map_basis10= 2ransac new
                \vspace{-0.45cm}
            \hspace{-3cm} 
        \end{subfigure}%                  
        \begin{subfigure}[b]{0.18\textwidth}
      %  \hspace{-3cm}
                \includegraphics[width=\textwidth]{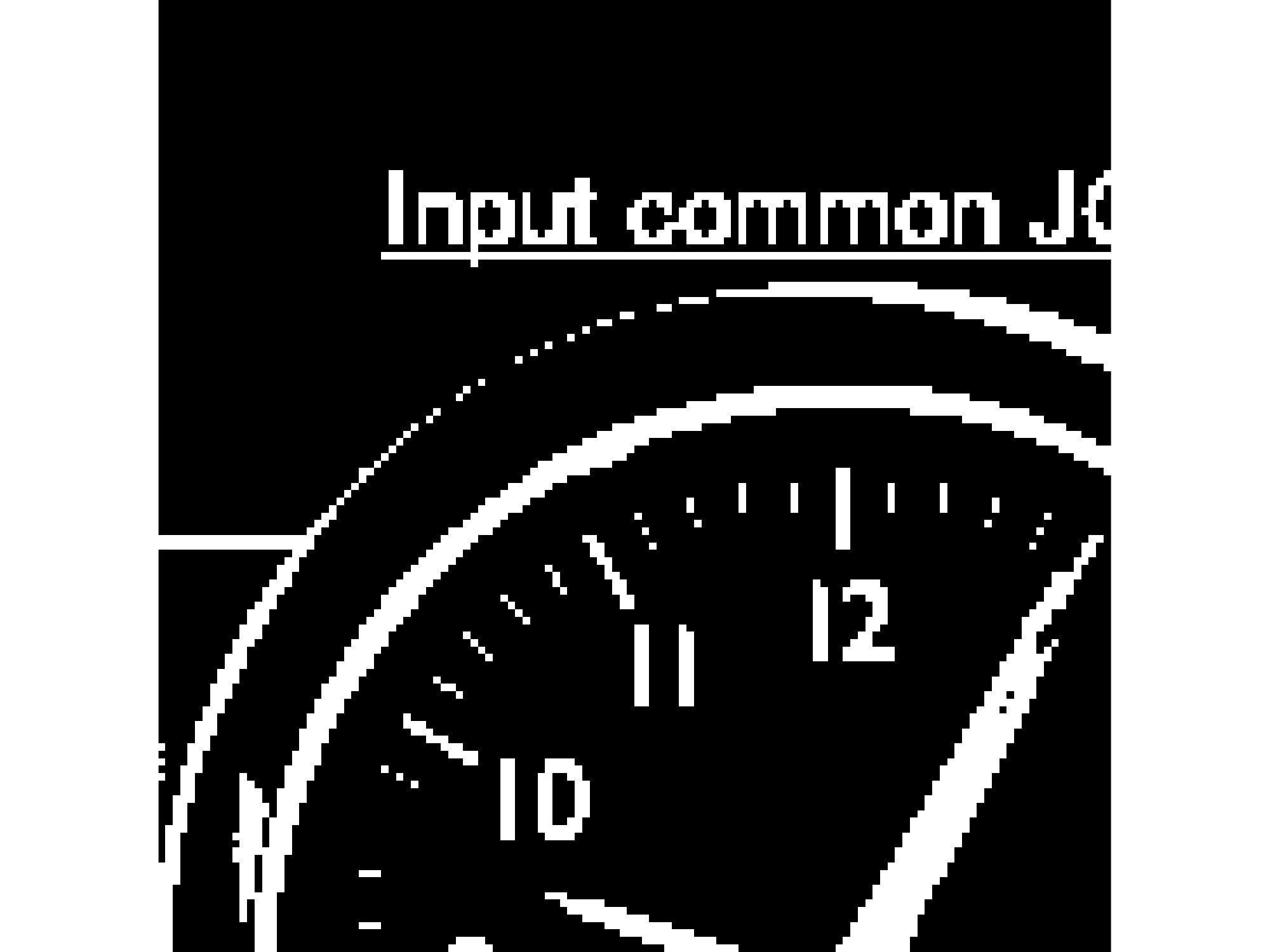}
                 \vspace{-0.45cm}
              \hspace{-4.8cm}
        \end{subfigure}
        \caption{Segmentation result for the selected test images. The images in the first row denotes the original images. And the images in the second, third, forth and the fifth rows denote the foreground map by shape primitive extraction and coding, hierarchical k-means clustering, least absolute deviation fitting and the proposed algorithm respectively.}
\end{figure*}

\begin{table}[h]
\centering
  \caption{Comparison of accuracy of different algorithms}
  \centering
\begin{tabular}{|m{3.4cm}|m{1.2cm}|m{1.2cm}|m{1.2cm}|}
\hline
Segmentation Algorithm  &  \  \ Precision & \ \  Recall & \  F1 score\\
\hline
SPEC \cite{spec} & \ \ \ 50\% & \ \ \  64\% & \ \ \ 56.1\% \\
\hline
 Hierarchical Clustering \cite{djvu} & \ \ \ 64\% & \ \ \ 69\% & \ \ \ 66.4\% \\
\hline 
 Least Absolute Deviation \cite{LAD} & \ \ \  91.4\% & \ \ \  87\% & \ \ \  89.1\% \\
\hline
 The proposed algorithm & \ \ \ 94.3\%  & \ \ \ 88\%  & \ \ \  90.9\%\\
\hline
\end{tabular}
\label{TblComp}
\vspace{-0.7cm}
\end{table}

It can be seen that in all cases the proposed algorithm gives superior performance over DjVu and SPEC.
There are also noticeable improvement over our prior approach using LAD.
Note that our dataset mainly consists of challenging images where the background and foreground have overlapping color ranges. For simpler cases where the background has a narrow color range that is quite different from the foreground, both DjVu and LAD will work well. On the other hand, SPEC  does not work well when the background is fairly homogeneous within a block and the foreground text/lines have varying colors.
The result for the rest of images in our dataset are publicly available and can be downloaded from \cite{our_dataset}.
We would like to discuss briefly about the impact of varying $K$, number of bases, on the accuracy of the algorithm. By increasing $K$, we will have more inliers, i.e. less foreground pixels. Therefore by increasing $K$ from its optimal value, we will get a higher precision and lower recall.

\section{Conclusion}
This paper proposed a new algorithms for segmentation of background and foreground in images. The background is defined as the smooth component of the image that can be well modeled by a set of DCT functions and foreground as a sparse component overlaid on background. We propose a sparse decomposition framework to decompose the image into these two layers.
Compared to our prior  least absolute fitting formulation, the background layer is allowed to choose as many bases from a rich set of smooth functions, but the coefficients are enforced to be sparse so that will not falsely include foreground pixels.
Total variation of the foreground component is also added to the cost function to promote the foreground pixels to be connected.
This algorithm has been tested on several test images and compared with three other well-known algorithms for background/foreground separation and has shown significantly better performance.

% use section* for acknowledgment
\section*{Acknowledgment}
The authors would like to thank Patrick Combettes and Ivan Selesnick for their useful comments and feedback regarding this work.
We would also like to thank JCT-VC group for providing the HEVC test sequences for screen content coding.

% Can use something like this to put references on a page
% by themselves when using endfloat and the captionsoff option.
\ifCLASSOPTIONcaptionsoff
  \newpage
\fi

%\begin{IEEEbiography}{Michael Shell}
%Biography text here.
%\end{IEEEbiography}
% if you will not have a photo at all:
%\begin{IEEEbiographynophoto}{John Doe}
%Biography text here.
%\end{IEEEbiographynophoto}

% insert where needed to balance the two columns on the last page with
% biographies
%\newpage
%\begin{IEEEbiographynophoto}{Jane Doe}
%Biography text here.
%\end{IEEEbiographynophoto}

\end{document}